\documentclass[11pt]{article}
\usepackage{shellesc}

\usepackage[]{acl}

\usepackage{times}
\usepackage{latexsym}
\usepackage{inconsolata}

\usepackage{times}
\usepackage{latexsym}
\usepackage{graphicx}
\usepackage{makecell}
\usepackage[subrefformat=parens]{subcaption}
\usepackage{caption}
\usepackage{subcaption}
\usepackage{booktabs}
\usepackage{multicol}
\usepackage{multirow}
\usepackage{microtype}
\usepackage{makecell}
\usepackage{pgfplots}
\usepackage{xcolor}
\usepackage{soul}
\usepackage{mathdesign}
\usepackage{textcomp}  % Required for encoding \textbigcircle
\usepackage{scalerel}  % Required for emoji \scalerel
\usepackage{epstopdf}

\newcommand{\ctext}[3][RGB]{%
  \begingroup
  \definecolor{hlcolor}{#1}{#2}\sethlcolor{hlcolor}%
  \hl{#3}%
  \endgroup
}

\def\headphones{\scalerel*{\includegraphics{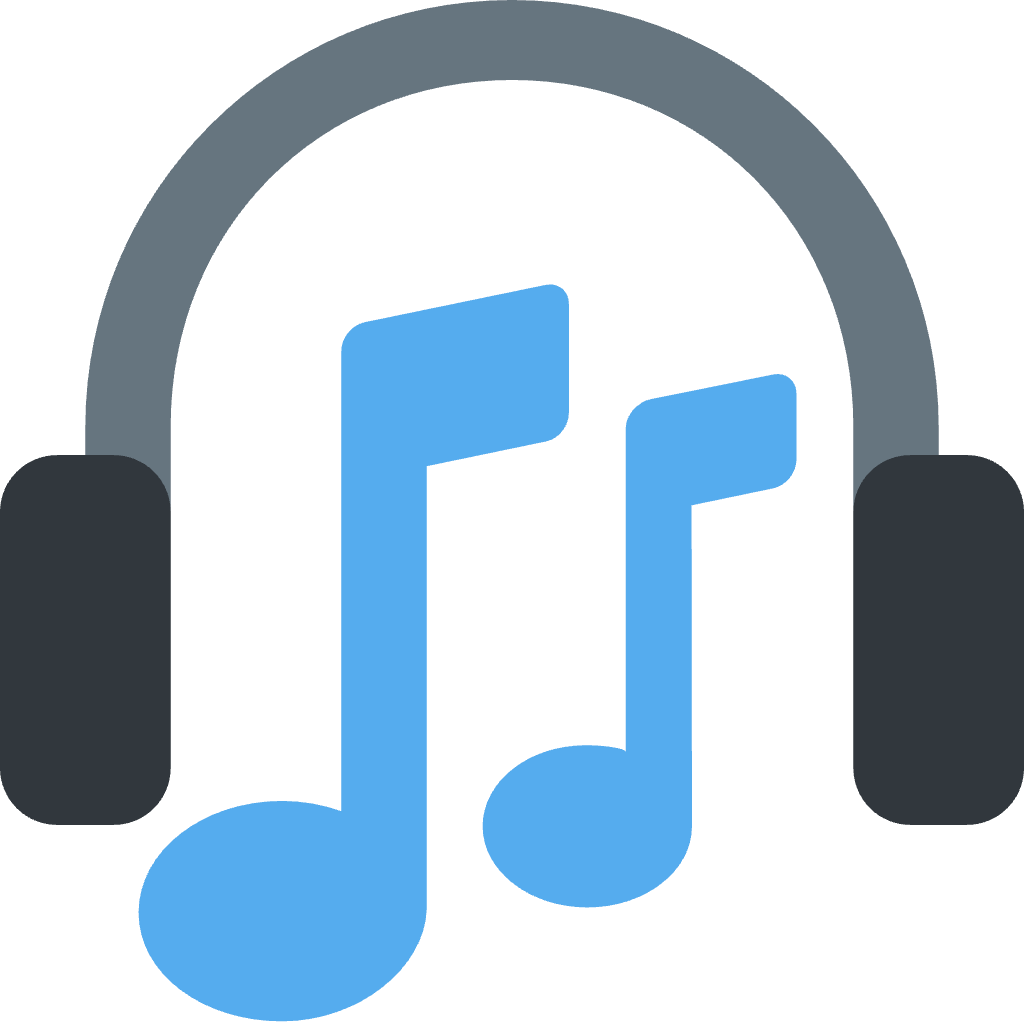}}{\textrm{\textbigcircle}}}

\def\hand{\scalerel*[10pt]{\includegraphics{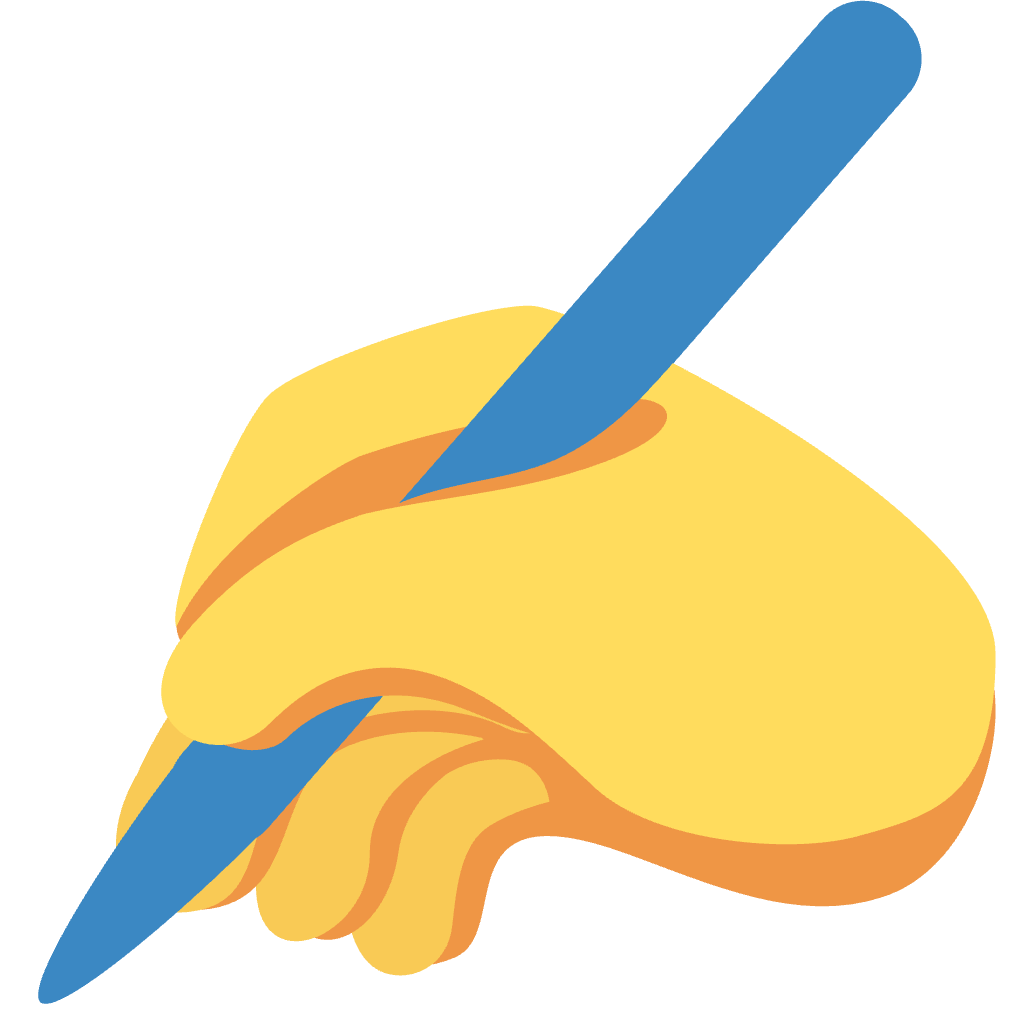}}{\textrm{\textbigcircle}}}

\def\magnifying{\scalerel*{\includegraphics{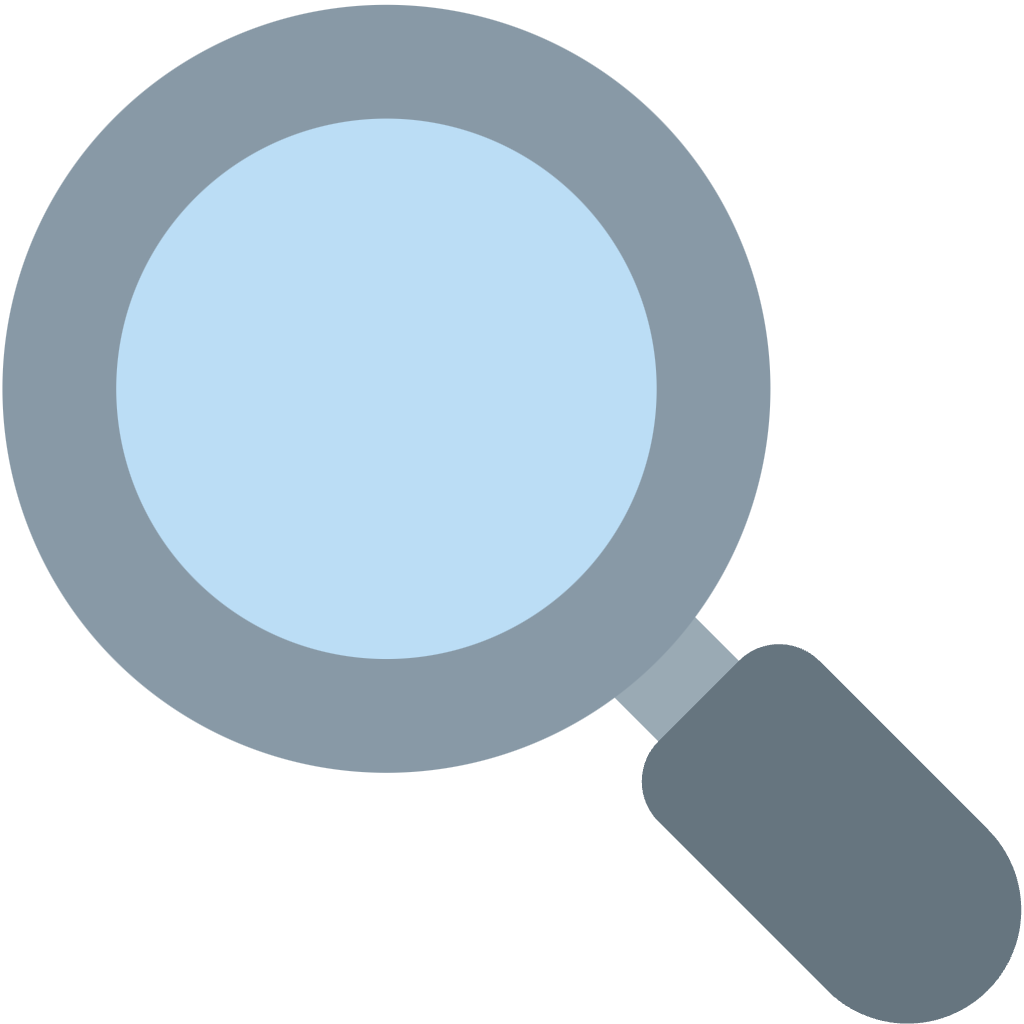}}{\textrm{\textbigcircle}}}

\def\chart{\scalerel*{\includegraphics{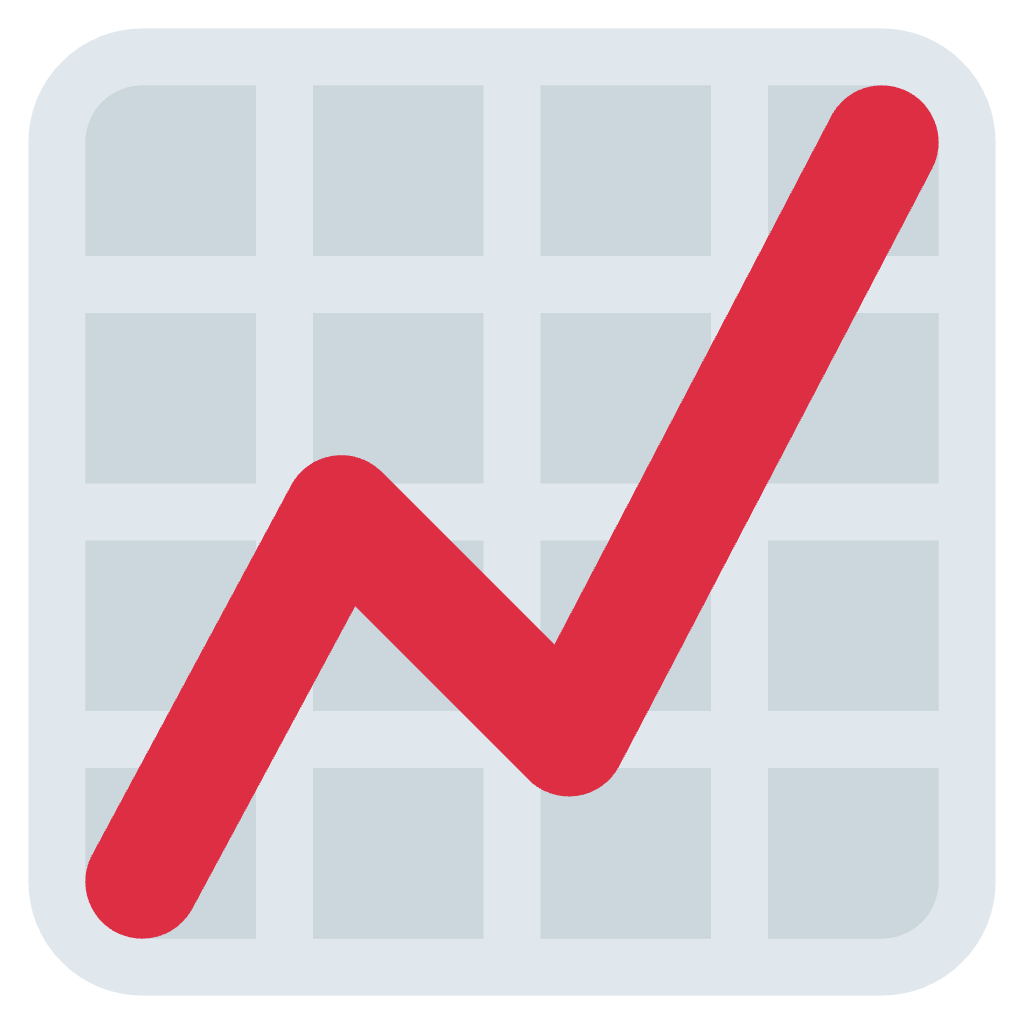}}{\textrm{\textbigcircle}}}

\def\clipboard{\scalerel*{\includegraphics{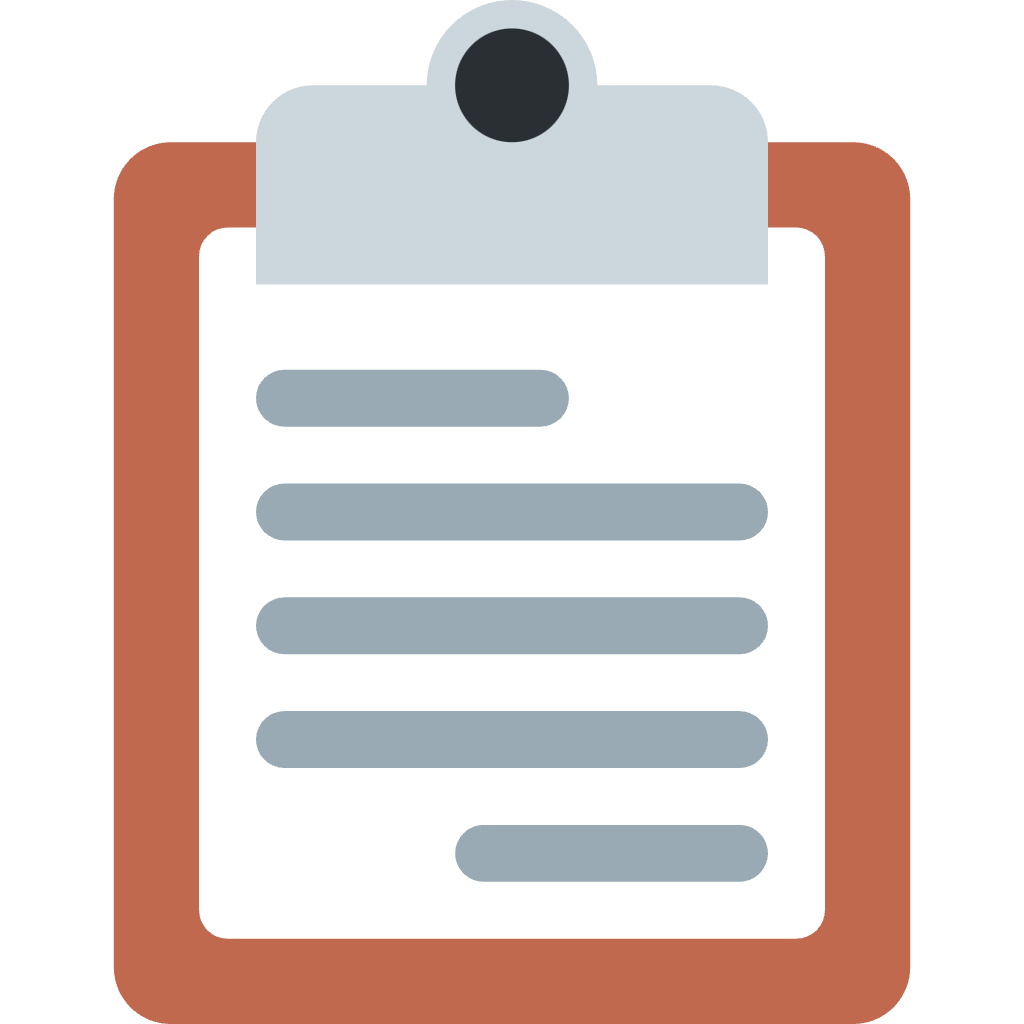}}{\textrm{\textbigcircle}}}

\usepackage[T1]{fontenc}
\usepackage[utf8]{inputenc}

\usepackage{microtype}

\usepackage{inconsolata}
\usepackage{amssymb}
\usepackage{amsmath}
\usepackage{multicol}

\usepackage{arydshln}

\newcolumntype{R}{>{\collectcell\ApplyGradient}c<{\endcollectcell}}
  
\title{From Facts to Insights: A Study on the Generation and Evaluation of Analytical Reports for Deciphering Earnings Calls}

\author{Tomas Goldsack,$^{1}$ Yang Wang,$^{1}$ Chenghua Lin,$^{1,2}$ Chung-Chi Chen$^{3}$  \\
        $^{1}$Department of Computer Science, University of Sheffield, UK \\
        $^{2}$Department of Computer Science, University of Manchester, UK \\ 
        $^{3}$Artificial Intelligence Research Center, AIST, Japan\\
        \small\texttt{\{tgoldsack1, Y.Wang4\}@sheffield.ac.uk} \quad \texttt{chenghua.lin@manchester.ac.uk} \\
        \small \texttt{c.c.chen@acm.org}}

\begin{document}
\maketitle
\begin{abstract}
% This paper conducts a comparative analysis of Journalistic and Analytical Reports derived from Earnings Calls (ECs), focusing on their readability, thematic content, and the effectiveness of Large Language Models (LLMs) in producing analytical reports. We examine the differences between these types of reports in terms of readability and subject matter. Addressing the gap in research on the generation of Analytical Reports from ECs, we investigate the capability of agent-based methods in creating and evaluating analytical reports based on ECs. Our findings suggest that involving more agents increases the \textcolor{orange}{complexity} of the generated reports. We also observe a significant correlation across most dimensions between GPT-4 and human evaluators. However, a noticeable difference emerges in the preference for reports authored by experts over those generated by agents, highlighting GPT-4's limitations in evaluating report quality. We conclude by suggesting several directions for future research aimed at producing high-quality, insightful analytical content for ECs.
This paper explores the use of Large Language Models (LLMs) in the generation and evaluation of analytical reports derived from Earnings Calls (ECs). 
% We conduct a comparative analysis of analytical reports with more commonly generated journalistic reports, focusing on key differences in style and thematic content. 
Addressing a current gap in research, we explore the generation of analytical reports with LLMs in a multi-agent framework, designing specialized agents that introduce diverse viewpoints and desirable topics of analysis into the report generation process. 
Through multiple analyses, we examine the alignment between generated and human-written reports and the impact of both individual and collective agents. Our findings suggest that the introduction of additional agents results in more insightful reports, although reports generated by human experts remain preferred in the majority of cases. 
Finally, we address the challenging issue of report evaluation, we examine the limitations and strengths of LLMs in assessing the quality of generated reports in different settings, revealing a significant correlation with human experts across multiple dimensions. 
% We conclude by outlining promising directions for future research aimed at producing high-quality, insightful analytical content for ECs. 
\end{abstract}

\section{Introduction} 
\label{sec:introduction}

Earnings Calls (ECs), critical quarterly meetings conducted by publicly traded companies to discuss financial performance with professional analysts, have been extensively studied for various prediction tasks. These tasks include volatility prediction \citep{sawhney-etal-2021-empirical, niu-etal-2023-kefvp}, analyst decision prediction \citep{keith-stent-2019-modeling}, financial risk prediction \citep{qin-yang-2019-say}, and earnings surprise prediction \citep{koval-etal-2023-forecasting}, highlighting ECs' significance in investment decision-making. Because ECs' typical duration of about one hour, another prominent research area in this domain 
%critical research area
is summarizing lengthy EC transcripts \citep{mukherjee-etal-2022-ectsum}. Post-EC, two types of summaries emerge: \textit{Journalistic Reports}, in which journalists concisely summarize the key financial takeaways from the meeting,
and \textit{Analytical Reports}, 
in which professional analysts offer a considerably more extensive and multifaceted analysis of meeting events, financial performance, and implications on investment strategies.
% \textcolor{red}{Please rewrite this part with this narrative: (i) prior work mostly focus on jornalistic report; (ii) analytics reports have been ignored/unexplored but have great values to [what audiences]].} 
Whilst the automatic generation of journalistic reports has been addressed in previous studies \citep{mukherjee-etal-2022-ectsum}, no work to our knowledge has explored the task of generating analytical reports, despite numerous potential benefits. For example, the automatic generation of analytical reports could reduce the burden placed on analysts, enable the immediate distribution of key information to a broad range of stakeholders, and introduce novel insights through scalable interpretation of complex data.
% \textcolor{orange}{reducing the burden placed on analysts (thus improving productivity); enhancing reporting factors such as consistency, accuracy, and customizability; and possibly introducing novel insights through the interpretation of complex data.}
% However, given its inherent complexity, this task remains a considerable challenge for a standard approaches.
However, given the inherent complexity of generating analytical reports, success in this challenging task requires a methodology that can enable an in-depth analysis across multiple varied and important technical aspects, such as expectations on future operations and managers' attitudes during ECs.

% This makes such an approach potentially well-suited to analytical report generation, which necessitates 
%While previous work has focused on journalistic reports \citep{mukherjee-etal-2022-ectsum}, 
% analytical reports and the key distinctions between these two report types remain largely unexplored. 
% Additionally, the automatic generation of \textcolor{red}{such reports} has the potential } 
% % \textcolor{red}{[CL: add some text to highlight why Analytical reports are important and useful.]}
% In this paper, we take the first step towards addressing this gap by conducting a pilot analysis comparing report types across various computational linguistics dimensions and, subsequently,  investigating the task of generating analytical reports. 
 
% The field of Natural Language Processing (NLP) has witnessed transformative changes in recent years, largely due to advancements in Large Language Models (LLMs). \textcolor{orange}{Breakthroughs such as instruction-tuning \citep{instruction-tuning} and Reinforcement Learning from Human Feedback (RLHF) \citep{Ouyang0JAWMZASR22} have enhanced LLMs' zero-shot performance and reasoning capabilities.} This, in turn, has given rise to remarkable role-playing abilities that have led to their widespread adoption in consumer-oriented platforms such as ChatGPT, whereby the underlying LLM (GPT-X) is effectively tasked with role-playing as a helpful assistant \citep{ShanahanMR23}.

One promising candidate for such an approach comes in the form of multi-agent frameworks: an exciting avenue of recent research that explores how multiple role-playing LLM ``Agents'' can be deployed to cooperatively solve a task. When deployed for generation tasks such as Software development \citep{qian2023communicative} and Trivia-based Creative Writing \citep{wang2024unleashing}, the introduction of role-based division has enabled complicated requirements to be broken down into simple subtasks and processes, reducing the cognitive and contextual burden placed the underlying models. Furthermore, the utilization of role-playing LLM agents has provided expanded opportunities for domain specialization, the leveraging of external data/tools, and the incorporation of diverse viewpoints, all of which could add significant value to the generation of analytical reports.
Notably, such an approach also bears a closer resemblance to a human writing process which, from a cognitive science perspective, is a complex, cyclic, and multi-step procedure, often requiring strategic discourse planning and multiple iterations to effectively achieve a communicative goal \citep{flower1981cognitive}.

In this work, we explore the utility of a multi-agent framework for generating analytical reports using LLMs.
% Utilizing a sample of reports collected from a professional source, we provide a thorough analysis of the various differences between analytical reports and journalistic reports (\S\ref{sec:report_types}).
% Subsequently,
We establish the details of our novel multi-agent framework and specially designed agents (\S\ref{sec:framework}), before performing a thorough characterization of the generated reports under different settings (\S\ref{sec:characterization}), highlighting key differences to human-authored reports and the additional insights offered by feedback agents. 
Finally, we assess the capabilities of 
LLMs in evaluating the quality of generated reports (\S\ref{sec:evaluation}), establishing both promising directions and limitations, and laying the groundwork for future research on this task. 

Overall, the following three research questions (RQ) are addressed:

\vspace{2mm}

\noindent \textbf{(RQ1)}: How do generated analytical reports differ from human-authored analytical reports? 

% \noindent \textbf{(RQ2)}: Can a multi-agent approach to generating analytical reports achieve superior quality compared to a single-agent approach?
\vspace{2mm}

\noindent \textbf{(RQ2)}: Can a multi-agent approach be used to generate more insightful analytical reports?

\vspace{2mm}

\noindent \textbf{(RQ3)}: How effective are LLM-based evaluation methods in assessing the quality of analytical reports? 
% generated by multi-agent systems?

% \begin{table}[t]
%     \centering  
%     \small
%     \begin{tabular}{lcccccccc}
%     \hline   
%      \textbf{Report} &&  \textbf{\# Sentences} & \textbf{FKGL} & \textbf{CLI} & \textbf{ARI}  \\ \hline
%         Journalistic && 252 & 5.73 & 8.78 & 7.64  \\ 
%         Analytical && 896 & 7.93 & 9.55 & 9.48  \\
%          \hline
%     \end{tabular}
%     \caption{Readability of journalistic and analytical reports.}
%     \label{tab:Readability}
% \end{table}

\begin{figure}[t]
    \centering
    \includegraphics[width=0.85\columnwidth]{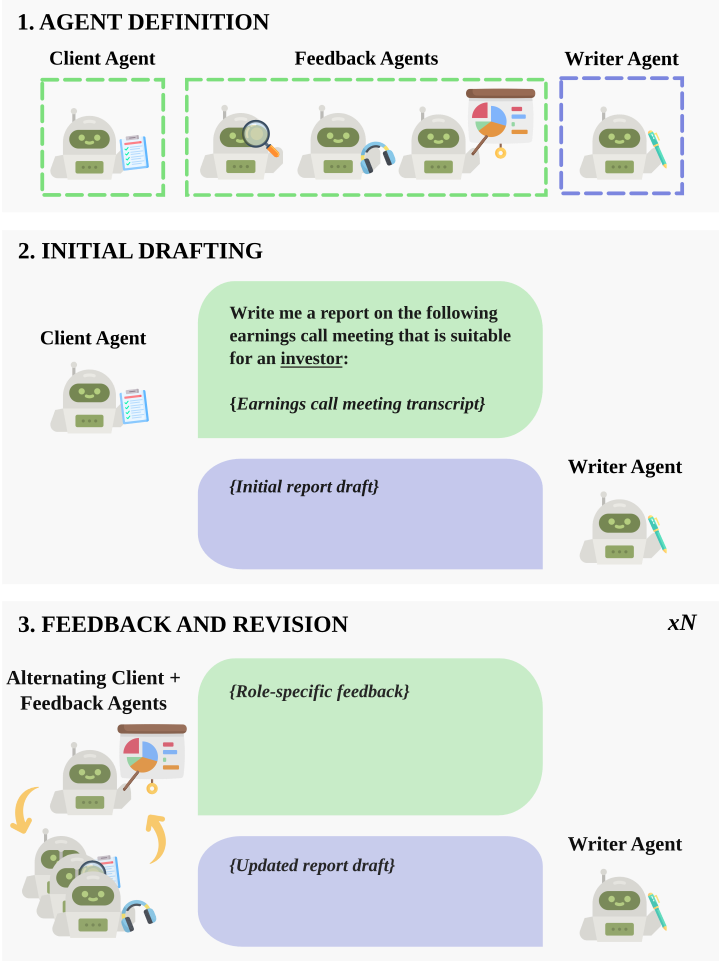}
    \caption{An overview of our multi-agent framework.}
    \label{fig:overview}
\end{figure}

\section{Multi-Agent Report Generation} \label{sec:framework}
We explore a novel framework for generating analytical reports through collaborative multi-agent conversation, leveraging the capacity of LLMs to refine their output based on feedback. This framework, developed using Microsoft's AutoGen \citep{wu2023autogen}, assigns each LLM-powered agent a distinct role through an initialization prompt, dictating their contribution to the conversation. For all experiments and agents, we employ the \texttt{gpt-4-1106-preview} model via the ChatGPT API as the underlying LLM.
Figure \ref{fig:overview} illustrates the framework, whereby a single agent with the role of  \textbf{Writer} (\hand) is tasked with drafting and revising the report, guided by feedback from other agents. The process encompasses three stages:

\paragraph{Stage 1. Agent Definition}
Before generation commences, it is imperative to define the agents involved: 1) a \textbf{Client} (\clipboard) agent, providing the initial task brief (as shown in Figure \ref{fig:overview}) and subsequent feedback representing the audience's perspective; 2) ``Feedback'' agents, offering insights based on their specific roles.
In this study, we explore the integration of three feedback agents alongside the Writer and Client agents. Details of the prompts used to initialize and generate responses for each agent are provided in Appendix \ref{sec:prompts}.

\vspace{3mm}

\noindent \textbf{Analyst} (\chart) agent tasked with extracting and analyzing historical financial data, the Analyst agent leverages the AlphaVantage API to gather earnings performance data for the previous quarter. This information, presented alongside the preceding conversation context to the LLM, allows the agent to draw additional insights about the company's current financial performance through comparison to previous performance, enabling the formulation of pertinent feedback.\footnote{\url{https://www.alphavantage.co/}}

\vspace{3mm}

\noindent \textbf{Psychologist} (\headphones) agent analyzes external data, specifically phonetic statistics from earnings call (EC) audio recordings, to offer additional insights on the level of confidence vocally expressed by management (e.g., CEO, CFO, etc.). Following \citet{qin-yang-2019-say} who show that such statistics are useful in the prediction of financial risk, PRAAT features are derived from the utterances of the management team during the EC, enriching the feedback provided to the LLM and the discourse on management's attitudes towards present or future financial performance.\footnote{Audio recordings are collected from Seeking Alpha (\url{https://seekingalpha.com/}) and force-aligned with transcripts using the Aenaes library (\url{https://github.com/readbeyond/aeneas}).}

\vspace{3mm}

\noindent \textbf{Editor} (\magnifying) agent ensures the generated report is suitable for the intended audience (in terms of content, style, and structure) and for checking that important information is maintained through revisions.

\paragraph{Stage 2. Initial Drafting}
Upon receiving the task brief from a Client agent, the Writer agent generates the initial draft of the report.

\paragraph{Stage 3. Feedback and Revision}
In an iterative process, each agent furnishes feedback aimed at enhancing the report, concentrating on elements relevant to their roles. Following each feedback round, the Writer updates the report. This cycle concludes when the preset maximum of \textit{N} iterations is reached or upon the Client agent's determination that the report is complete.\footnote{We set the value of N to 10 for all experiments, but find that it rarely reaches this threshold without being stopped by the Client.}

\section{Generated Report Characterization} \label{sec:characterization}

To gain a full understanding of the style, content, and utility of generated reports, we conduct several in-depth experiments and analyses. 
Adopting multiple configurations of our multi-agent framework, we generate reports for a sample of 60 EC transcripts used in previous work \citep{mukherjee-etal-2022-ectsum}, basing our sample size on previous work for multi-agent text generation tasks~\citep{chan2023chateval,wang2024unleashing}.%}

%\paragraph{How do generated reports differ from human-authored reports?}
\subsection{Generated vs. Human-Authored Reports} \label{subsec:generatedvshuman}
Given that our analytical reports are generated in a zero-shot setting, they is no guarantee that they will closely resemble those written by human experts.
Therefore, to answer \textbf{RQ1}, we start by analyzing the similarities and differences between generated reports and those produced by human experts. For human-authored reports, we use a sample of 26 equality research reports from the Bloomberg Terminal that are authored by professional analysts at J.P. Morgan, to which we were granted restricted access.
% These reports aim to provide an in-depth analysis of the company's financial performance for internal stakeholders and investors.

\begin{figure}[t]
    \centering
    \includegraphics[width=\columnwidth]{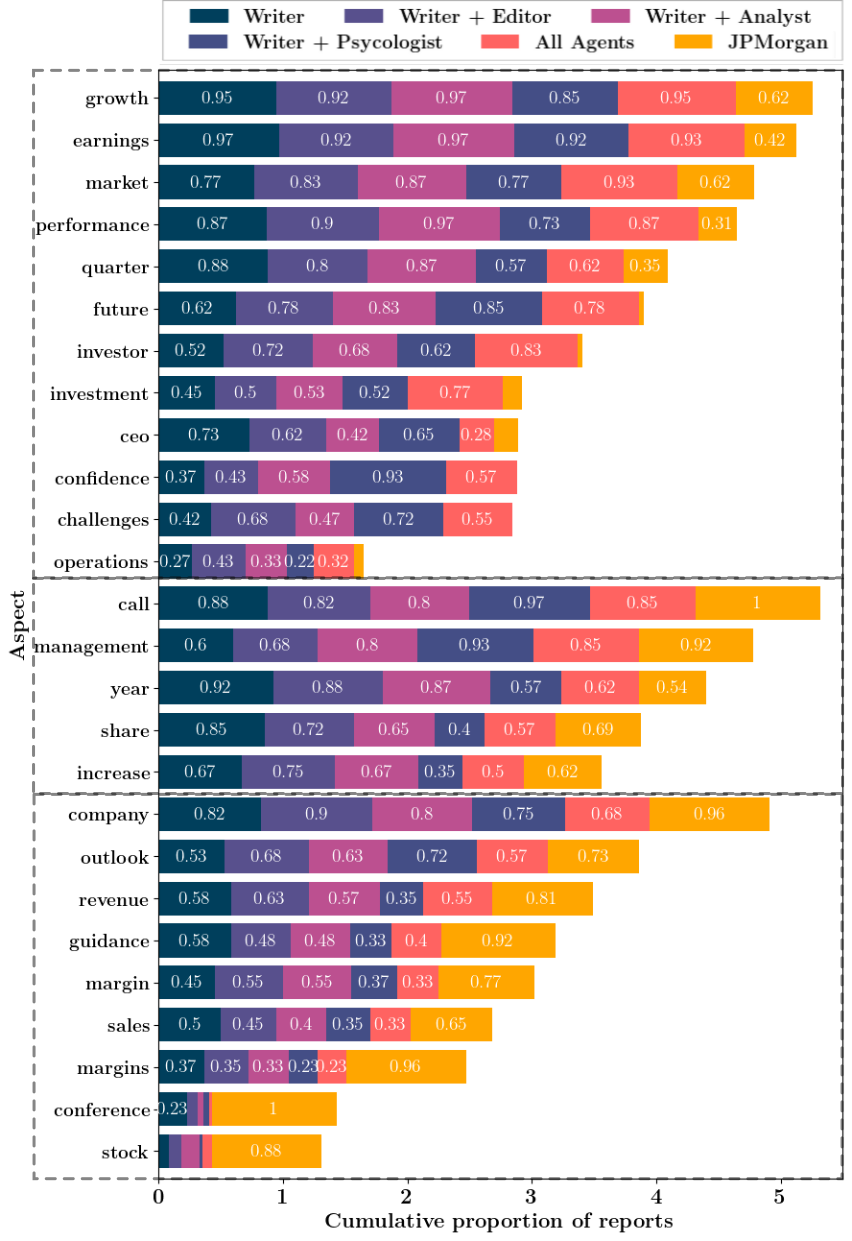}
    \caption{A visualization of the proportion of aspect occurrences for different report types, including at least the 10 most common aspects for each.}
    \label{fig:aspect_graph}
\end{figure}

\paragraph{Content} To identify the key topics of discussion within each report type, we employ the aspect-extraction method outlined by \citet{tulkens-van-cranenburgh-2020-embarrassingly}, we use SpaCy \citep{spacy2} to extract the 250 most frequent nouns in the earnings call transcripts of ECTSumm and analyze their presence in both generated and reference (JP Morgan) reports. 
Figure \ref{fig:aspect_graph} presents the results of the content analysis.
% a content analysis that illustrates the occurrence of different topics in reports generated under various settings, as well as the reports of JP Morgan. 
Using the Figure we can see that, for example, the topic of ``growth''%" 
appears in 95\% of analytical reports generated with all agents (All Agents), but only in 65\% of the human-authored reports of JP Morgan.
To enhance clarity, we've divided the figure into three segments based on the similarity of aspect occurrences between the generated reports (with all agents) and the reference reports. The top segment represents aspects that occur more frequently in the generated reports, the middle segment includes aspects with similar frequencies, and the bottom segment comprises aspects that occur more frequently in the reference reports. The Figure shows that, although generated and human-authored reports can be seen to discuss aspects like ``share'', ``management'', and ``increase'' at a similar rate, there exists a significant divergence in content emphasis.
For instance, human-authored reports place a greater focus on financial statistics, with aspects like ``margin(s)'', ``revenue'', ``sales'', and ``stock'' occurring more frequently.  
In contrast, generated reports can be seen to emphasize aspects such as ``performance'', ``future'', ``earnings'', and ``market'' which, whilst different from references, remain indicative of analytical discussion. 
Furthermore, generated reports introduce several aspects that very rarely occur in reference reports (or do not occur at all), including ``investor'', `investment'', and ``confidence'' implying they address their intended audience more explicitly.

\paragraph{Style} Table~\ref{tab:Readability with different numbers of feedback agents} presents the statistical results of several metrics selected to measure the stylistic properties of reports.
Here, we employ readability metrics Flesch-Kincaid Grade Level (FKGL)~\cite{kincaid1975derivation}, Coleman–Liau index (CLI)~\cite{coleman1975computer}, and Automated Readability Index (ARI)~\cite{senter1967automated}, to assess text complexity~\cite{goldsack-etal-2023-enhancing}. These widely used metrics are calculated based on such factors as the number of characters (ARI and CLI), syllables (FKGL), words (all), and sentences (all) present in a given text. Higher scores indicate greater document complexity. 
% For instance, the FKGL is applied to assess business policies and financial documents across numerous U.S. states. 
To provide insights on document length and content novelty, we also calculate the number of sentences using NLTK \citep{bird2009natural} and the abstractiveness (\% of unigrams used that do not occur in the source transcript).

Here, we can see that, although both generated and human-authored reports are generally of a similar length and level of abstractiveness, there is a significant divergence in the scores of readability metrics. Expert-written reports obtain readability scores ranging from 7-10, deemed suitable for a majority of marketing materials, and indicative of shorter statement-like sentences (i.e., containing fewer syllables or characters). Contrastingly, readability scores of generated reports span a range of 12-20, indicative of longer more complex sentences and aligning with materials intended for a highly skilled readership, such as academic publications.
% \footnote{Note these observations are consistent with reference-less metric scores for JPMorgan instances, which are included in Table \ref{tab:jpmorgan_referenceless} in the Appendix.}

% \textcolor{red}{[CL: highlight num resutls.]}

% Figure \ref{fig:aspect_graph} visualizes the most discussed aspects in reports generated under different settings, as well as the references of ECTSumm and JPMorgan.
% We utilize this Figure alongside Table~\ref{tab:Readability with different numbers of feedback agents} presents the statistical results of readability scores to identify how the content of the reports change under configurations that include the respective agent compared with those generated using only the Writer agent.  

\begin{table}[t]
  \centering
    \resizebox{\columnwidth}{!}{
    \begin{tabular}{lccccc}
    \hline
     \textbf{Agents} & \textbf{\# Sents} & \textbf{FKGL} & \textbf{CLI} & \textbf{ARI} & \textbf{Abst} \\
     \hline
     \hand & 24.35 & 12.88 & 16.42  & 16.87  & 41.74 \\
     \hand\magnifying & 22.90 &  13.67 & 17.55 & 17.83 & 48.03 \\
    \hand\chart & 21.43 & 13.44 & 17.32 & 17.24 & 49.46 \\
    \hand\headphones & 20.03 & 15.71 & 19.03 & 20.26 & 57.95 \\
    \hand\chart\magnifying & 19.65 & 14.76 & 18.33 & 19.10 & 53.40 \\
    \hand\headphones\magnifying & 19.68 & 15.69 & 19.18 & 20.11 & 56.87 \\
    \hand\headphones\chart\magnifying & 18.58 & 15.11 & 18.98 & 19.46 & 56.72 \\
    \hdashline
    % References & 37.33 & 7.93 & 9.55 & 9.48 & 52.45 \\  \hline
    References & 19.25 & 7.26 & 8.54 & 8.85 & 47.14 \\ \hline
    \end{tabular}%
    }
  \caption{Readability with different feedback agent configurations.}
  \label{tab:Readability with different numbers of feedback agents}%
\end{table}%

%\paragraph{The Impact of Feedback Agents}
\subsection{Impact of Feedback Agents}

% Editor
Again utilizing Figure \ref{fig:aspect_graph} and Table~\ref{tab:Readability with different numbers of feedback agents}, we can begin to assess the impact each agent has on the output report. 
% \textcolor{orange}{shows a reduction in the use of generally related terms} (e.g., ``earnings'', ``call'', ``quarter'', ``end'', ``share'') and
Firstly, the incorporation of both the Editor and the Analyst agents is shown to have a similar effect on content, increasing the rate at which aspects like ``outlook'', ``market'', ``management'', and ``future'' are discussed when compared to the Writer agent alone. 
This implies that both the additional financial statistics introduced by the Analyst and the critical feedback of the Editor induce a broader and more speculative analysis of company performance, causing the content to transition from focusing primarily on reporting the facts and figures from the transcript to a more speculative and potentially more insightful discussion. Table \ref{tab:Readability with different numbers of feedback agents} supports this, showing an increase in abstractiveness (Abst) upon the introduction of each agent, suggesting that report content becomes less based on the source transcript and more based on external data and agent discussion.  
Additionally, readability metrics (FKGL, CLI, and ARI) can also be seen to increase, denoting the use of longer and more complex words/sentences. 

% Psycologist
For the Psychologist agent, Figure \ref{fig:aspect_graph}, shows a significant decrease in the reporting of aspects relating to financial performance figures (``year'', ``share'', ``quarter'', ``increase'', ``revenue'', ``margins'') in favor of aspects relating to the attitude and confidence of management (``management'', ``confidence'', ``future'', ``outlook''), areas that this agent was designed to focus on. 
In addition to the change of focus, Table \ref{tab:Readability with different numbers of feedback agents} suggests a significant change in the style of reporting. More than any other agent, the Psychologist causes readability and abstractiveness metrics to rise, demonstrating its ability to influence the generated text through the introduction of novel content.

\begin{table*}[]
    \centering
    \resizebox{0.8\textwidth}{!}{%
    \begin{tabular}{lm{0.70\textwidth}}
       \hline
       \textbf{Report characteristic} & \textbf{Description} \\
       \hline
       Financial takeaways  & The key financial details from the meeting (i.e., numerical statistics relating to company performance for the quarter).  \\ \hline
       Financial context & Any additional information (e.g., financial details from previous quarters) that helps to contextualize the current financial performance. \\ \hline
       Management attitudes & Information on how management (e.g., CEO, CFO, etc..) feels about the company’s financial performance. \\ \hline
       Management expectation  &  Details about how the company is expected to perform in the future/next quarter. \\ \hline
%        Noteworthy meeting events & 
% Any details about noteworthy events that could have occurred in the meeting that might be relevant to an investor (e.g., repeated questioning from analysts on a particular topic, etc.). \\ \hline
      Possible future events &
Details surrounding any noteworthy events/scenarios that are likely to occur in the future. \\ \hline
    \end{tabular}
    }
    \caption{Human evaluation assessment criteria descriptions.}
    \label{tab:human_eval-aspects}
\end{table*}

%\paragraph{How Insightful are Generated Reports?}
\subsection{Insightfulness of Generated Reports}\label{subsec:insightfulness}
To answer \textbf{RQ2}, and determine how effective a multi-agent approach is at providing insights that are potentially useful to an investor, we conduct an in-depth human evaluation utilizing domain experts. We employ three evaluators, all of whom are pursuing postgraduate studies in Finance, and ask them to assess reports generated by the Writer (\hand) alone with those produced using all agents (\hand\headphones\chart\magnifying) for 32 randomly-selected earnings call instances, allowing us to see the impact of our designed feedback agents. Specifically, the evaluators' task is broken down into the assessment of the following key report characteristics: 1) Financial takeaways, 2) Financial context, 3) Management attitudes, 4) Management expectations, 
% 5) Notablworthy meeting events, 
5) Possible future events. 
For each characteristic, evaluators assign one of the following labels, identifying both the presence and insightfulness of relevant information: 1) Reported and insightful, 2) Reported and reasonable, 3) Reported but not useful, and 4) Not reported. Our evaluation characteristics and labels were developed in collaboration with professional analysts, and are crucial in capturing report value to potential investors. Full descriptions of our characteristics and labels are provided in Tables \ref{tab:human_eval-aspects} and \ref{tab:human_eval-labels} (Appendix \ref{sec:additional_experiments}), respectively.\footnote{A discussion of inter-annotator agreement is also provided in Appendix \ref{sec:additional_experiments}.}

\begin{figure}[t]
    \centering
    \resizebox{0.85\columnwidth}{!}{%
    \includegraphics{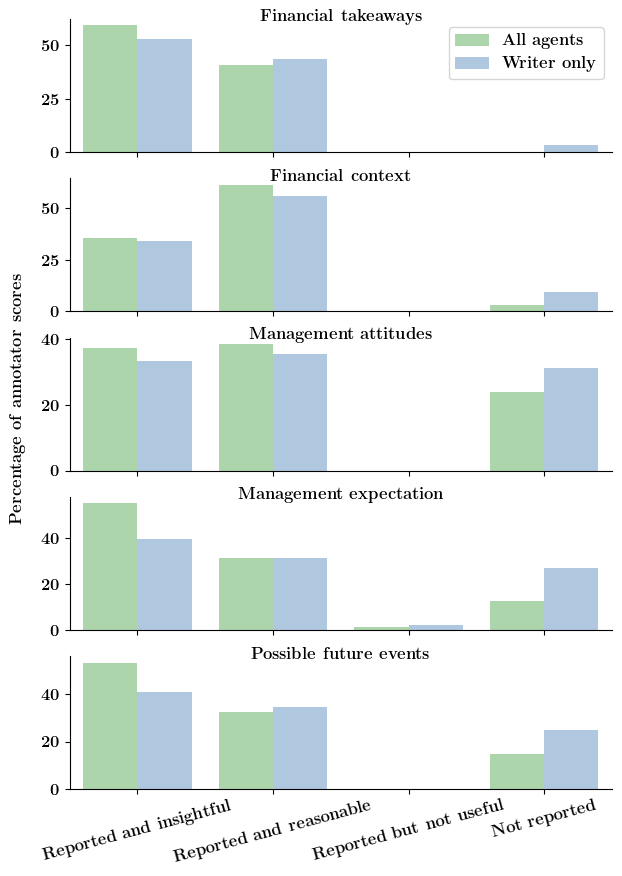}}
    \caption{Characteristic-based human evaluation results.}
    \label{fig:human_eval}
\end{figure}

\paragraph{Evaluation results}
The distribution of evaluator ratings for each system is depicted in Figure~\ref{fig:human_eval}. 
Firstly, it can be observed that, for both systems, the percentage of positive labels (i.e., ``reported and insightful" and ``reported and useful") significantly outweighs the more negative labels for all characteristics. 
Additionally, there are almost no cases of the "reported but not useful" label for either system, attesting to the quality of the information that is included in the report. Of all the characteristics that are discussed in the reports generated by both systems, Financial takeaways are the most frequently included, whereas Management attitudes are the most frequently absent.  

Results also indicate that the inclusion of all agents leads to a decrease in the percentage of characteristics labeled as ``not reported'' and an increase in the percentage of characteristics deemed ``reported and insightful'' in all instances. Particularly notable improvements are observed for the management expectations and possible future events characteristics, illustrating how the additional feedback agents significantly improve the speculation and anticipation of future performance in generated reports, factors that are particularly relevant to investment decision-making.

\begin{figure*}[t]
    \fbox{%
        \parbox{0.955\textwidth}{%
            \begingroup
                \fontsize{8pt}{10pt}\selectfont
                \textbf{A. Financial takeaways}
                    
                \fontsize{8pt}{10pt}\selectfont   
                \textbf{Writer only} - \ctext[RGB]{255,213,188}{
Motorola Solutions showcased a robust financial performance in the third quarter of 2021. The company reported a revenue increase of 13\% and a surge in earnings per share (EPS) by 21\%, signaling strong demand across businesses. Operating margins also expanded by 150 basis points. A record backlog of \$11.4 billion was achieved, marking a 7\% year-over-year increase.}\boldmath{$^{1/3}$}

                \fontsize{8pt}{10pt}\selectfont    
                \textbf{All agents} - \ctext[RGB]{189,231,199}{Motorola Solutions persists in its upward financial trajectory, as illustrated by pivotal metrics:

- \textit{Revenues}: A robust 13\% increase substantiates the company's firm market footing and demand for its solutions.
- \textit{Earnings Per Share (EPS)}: The 21\% rise in EPS reflects burgeoning profitability, a reassuring signal for shareholders.
- \textit{Operating Margins}: The expansion by 150 basis points testifies to enhanced operational rigor and disciplined cost governance.
- \textit{Record Backlog}: The \$11.4 billion figure – up 7\% year-over-year – fortifies visibility into future revenue streams.}\boldmath{$^{3/3}$}
                
            \endgroup \vspace{-4pt} 
            \rule{0.96\textwidth}{0.1pt} \\
            \begingroup
                \fontsize{8pt}{10pt}\selectfont
                \textbf{b. Management attitudes}

                \fontsize{8pt}{10pt}\selectfont
                \textbf{Writer only} - \ctext[RGB]{255,213,188}{As the economy continues to transition, Dynex remains cautiously optimistic about capitalizing on emerging opportunities while maintaining its established principles of responsible money management and transparency.}\boldmath{$^{1/3}$}

                \fontsize{8pt}{10pt}\selectfont
                \fontsize{8pt}{10pt}\selectfont    
                \textbf{All agents} - \ctext[RGB]{189,231,199}{The managers' communication styles, marked by confidence and clarity, supplemented the strong financial results by providing an indication of management's collective poise and tactful approach to stewardship, which might positively influence investor sentiment.}\boldmath{$^{3/3}$}
                
            \endgroup \vspace{-4pt} 
            \rule{0.96\textwidth}{0.1pt} \\
            \begingroup
                \fontsize{8pt}{10pt}\selectfont
                \textbf{c. Management expectation}

                \fontsize{8pt}{10pt}\selectfont
                \textbf{Writer only}  -  \ctext[RGB]{255,213,188}{In regards to the outlook for Q4 of fiscal '22, RPM projected consolidated sales growth in the low teens versus the prior year, despite anticipated supply chain challenges and currency headwinds due to the strengthening U.S. dollar. Notwithstanding these obstacles, it was suggested that RPM could achieve record EBIT on a consolidated basis in Q4.}\boldmath{$^{1/3}$}

                \fontsize{8pt}{10pt}\selectfont
                \textbf{All agents} - \ctext[RGB]{189,231,199}{\textit{Growth Prospects}: RPM foresees low teens growth in consolidated sales and adjusted EBIT for Q4. This would continue the trend of sustained growth even when set against a challenging economic backdrop, including impacts from the Ukrainian conflict.}\boldmath{$^{3/3}$}
                
            \endgroup 
        }
    } % elife-69719-v2
    \caption{A case study comparing how the introduction of feedback agents changes the discussion of key aspects. Colors and superscript are used to denote the number of evaluators who judged the sentence as ``reported and insightful (e.g., \boldmath{$^{3/3}$} = 3 out of 3 evaluators).}
    \label{fig:case_study}
\end{figure*}

\paragraph{Case study}
To gain a better understanding of insights generated by the multi-agent framework, we perform a closer examination of reports and annotator classifications, presenting examples of our findings in the form of a case study in Figure \ref{fig:case_study}.
The given examples demonstrate how additional agent feedback can make the reporting improve the reporting of important information. Looking at each of the provided instances, example A shows how the same financial takeaways are presented alongside additional insights for investors (a likely influence of the Editor); example B illustrates how novel content relating to the confidence of management is introduced by the Psychologist's audio analysis; and example C demonstrates how past financial from the Analyst can contextualize relevant facts.

% From this, we can infer that automatic reference-based metrics are unable to fully capture the quality of the generated report, particularly in scenarios when there is a deviation in style between prediction and reference.

\section{Evaluating Analytical Reports with LLMs} \label{sec:evaluation}
In this section, we address \textbf{RQ3} and attempt to establish how future works may address the challenge of evaluating generated analytical reports. Specifically, we first describe the challenges of a conventional reference-based evaluation (\S\ref{subsec:eval_challenges}), before exploring the potential of LLMs for reference-free evaluation (\S\ref{subsec:llm_eval}).

\subsection{Challenges of Reference-based Evaluation}
\label{subsec:eval_challenges}
Significantly, the more conventional method of assessing generated reports against references using automatic metrics faces several inherent limitations in the context of analytical reports. Firstly, the scarcity of available data samples is compounded by the fact that these reports, being internally generated by corporate entities, pose challenges in establishing public benchmarks. The small dataset of 26 reports we've gathered is subject to strict redistribution restrictions, precluding us from making them publicly accessible for benchmark creation.
Even if such access were feasible, these reports typically adhere to in-house guidelines and 
%contain internal information that is inaccessible,
practices, contributing to disparities in content and style between human and machine-generated reports such as those outlined in \S\ref{subsec:generatedvshuman}. Accordingly, any novel insights provided during generation are unlikely to be adequately captured or rewarded through reference-based comparison.  
Furthermore, these instances are based on Earnings Calls (ECs) considerably older (2012-2016) than others utilized in this study (2019-2022), which played a role in our granted access. This raises questions about their usefulness as a potential point of comparison, considering a possible evolution of financial reporting practices.
% \footnote{We share the specific earnings call meetings from which they are derived in Table \ref{tab:jpmporgan_ecs} in the Appendix.} 

\begin{table*}[t]
  \centering
  \small
    \begin{tabular}{lcccccccccccc}
          % & \multicolumn{3}{c|}{Financial Takeaways} & \multicolumn{3}{c|}{Financial Context} & \multicolumn{3}{c|}{Management Attitudes} & \multicolumn{3}{c}{Management Expectation} \\
         \hline
         \multirow{2}{*}{\textbf{Characteristic}} && \multicolumn{3}{c}{\textbf{GPT-4}} && \multicolumn{3}{c}{\textbf{Gemini-pro}} && \multicolumn{3}{c}{\textbf{Mistral-medium}} \\
         \cline{2-5} \cline{7-9} \cline{11-13}
          && \multicolumn{1}{c}{$\gamma$} & \multicolumn{1}{c}{$\rho$} & \multicolumn{1}{c}{$\tau$} && \multicolumn{1}{c}{$\gamma$} & \multicolumn{1}{c}{$\rho$} & \multicolumn{1}{c}{$\tau$} && \multicolumn{1}{c}{$\gamma$} & \multicolumn{1}{c}{$\rho$} & \multicolumn{1}{c}{$\tau$} \\ 
    \hline
    Financial Takeaways    && 0.375 & 0.160  & 0.412 && 0.156 & 0.018 & 0.014 && 0.139 & 0.205 & 0.192 \\
    Financial Context      && 0.597  & 0.455  & 0.397   && 0.341  & 0.330  & 0.292  && 0.758  & 0.437  & 0.397 \\
    Management Attitudes   && 0.570  & 0.524  & 0.463   && 0.248  & 0.301  & 0.266  && 0.463  & 0.558  & 0.492\\
    Management Expectation && 0.529  & 0.511  & 0.441   && 0.643  & 0.598  & 0.521  && 0.670  & 0.661  & 0.581 \\ 
    Future Events          && 0.472  & 0.379  & 0.327   && 0.179  & 0.194  & 0.167  && 0.422  & 0.382  & 0.330 \\ \hline
    Average                && 0.509 & 0.405 & 0.408   && 0.313  &  0.288  & 0.252  && 0.490  & 0.449  & 0.398 \\
    \hline
    \end{tabular}%
    \caption{Correlation statistics of LLMs vs. human evaluators (averaged) for each report characteristic.}
  \label{tab:GPT-4 vs. human evaluators}%
\end{table*}%

\subsection{Reference-free Evaluation with LLMs}
\label{subsec:llm_eval}

% \paragraph{Using LLMs as Evaluators}

Given the limitations described in \S\ref{subsec:eval_challenges}, we explore the use of LLMs for the reference-free evaluation of generated outputs, a direction that has proved promising in previous studies \citep{liu-etal-2023-g, luo2023chatgpt,chan2023chateval}. Utilizing their respective APIs, we experiment with GPT-4 \citep{openai2023gpt4}, Gemini-pro \citep{geminiteam2023gemini}, and Mistral-medium \citep{jiang2024mixtral}, instructing each to embody a financial expert and reenact evaluations performed by experts. We assess the performance of LLM evaluators in two popular human evaluation settings: 1) a characteristic-based setting and 2) a preference-based (ranking) setting. The prompts used for each setting are provided in Appendix \ref{sec:prompts}, Table \ref{tab:evaluator_prompts}. 

\paragraph{Characteristic-based Evaluation}
For a characteristic-based evaluation, LLM evaluators are tasked with performing the evaluation described in \S\ref{subsec:insightfulness}. Here, we adhere to established evaluation methodologies from previous studies~\cite{zhong-etal-2022-towards,chan2023chateval} and employ Pearson ($\gamma$), Spearman ($\rho$), and Kendall ($\tau$) correlation coefficients between LLM and human evaluators.\footnote{To calculate correlation, we convert our labels into numeric scores ranging from 1-4, with 4 being the most positive classification (reported and insightful) and 1 being the most negative (not reported).} 
Table~\ref{tab:GPT-4 vs. human evaluators} presents our findings.
Here, we can see that GPT-4 and Mistral obtain a good level of correlation with human experts, whereas Gemini-pro performs slightly works in terms of average correlation scores.
% Mistral achieves the highest average correlation.

Looking closer at individual characteristics, all models can be seen to achieve a strong level of correlation (> 0.5) for at least one of the listed characteristics and GPT-4 maintains at least a moderate level (> 0.3) of correlation across all characteristics. Contrastingly, Gemini and Mistral achieve a broader range of correlation scores, with particularly strong scores for some aspects (e.g., Management Expectations), but weaker scores for others (e.g., Financial Takeaways). 

Overall, these results indicate that LLMs have significant potential in the evaluation of analytical reports when assessing fine-grained characteristics, but that performance is likely to differ depending on the LLM. 
Although GPT-4 can be considered the best all-round evaluator, the fact that different LLMs achieve stronger correlations for specific characteristics is something that future works should consider when designing their evaluation.

% Looking closer at individual characteristics, we can see that all LLMs achieve relatively good alignment with evaluators' average scores in four out of five dimensions. For the remaining characteristic, key financial takeaways, 
% models can be seen to obtain lower (and often negative) correlation scores across all metrics. Seeking an explanation for this result, we examine the criterion-based correlation between annotators (Table \ref{tab:human evaluators vs. human evaluators} in the Appendix). Here, we find high disagreement for this characteristic, possibly arising out of differing standards when it comes to identifying which financial figures are ``key''. 
% % and which meeting events are ``noteworthy''. 
% However, for this instance, LLMs still exhibit a good correlation with at least one of the annotators (Table \ref{tab:GPT-4 vs. human evaluators - all} in the Appendix), suggesting some alignment in terms of these standards.

\paragraph{Preference-based Evaluation}
In addition to the characteristic-based evaluation, we perform a preference-based evaluation utilizing the professional analytics reports we collect from JP Morgan (described in \S\ref{subsec:generatedvshuman}). Specifically, evaluators are required to compare the reference report and the report generated by our system with all possible agents (\hand\chart\magnifying)\footnote{Note that, due to these ECs being from earlier dates, their audio recordings are unavailable and we were unable to include our Psychologist agent in report generation process for these instances.}
Here, we integrate argument quality evaluation principles~\cite{gretz2020large}, which hinge on \textit{whether evaluators would recommend a friend to use that argument as is in a speech supporting/contesting the topic, regardless of personal opinion}.
In our case, evaluators indicate which report they would recommend to someone who would be making an investment decision based on the information released in the EC.
% Although this comparison represents a somewhat challenging benchmark given the access to internal company data, the results are likely to prove insightful in providing a sense of how close to (or far away from) human expert performance our system can achieve. 
% Being a largely subjective decision, we present the average preference of each evaluator (alongside the overall average) in Table \ref{tab:jpmorgan_human_eval}. 

After conducting this evaluation using the same three experts as in our characteristic-based evaluation, we find that there remains a large preference for human-authored reports over generated reports, with human annotators preferring the reference reports of JP Morgan  83.33\% of the time (on average). 
% with the individual preference of annotators ranging from 100\% of reports to 58.33\%. However, it should be noted that for even one annotator to consider generated reports preferable in 41.67\% of cases is an interesting result, highlighting the potential usefulness of such generated reports when expert-written reports are unavailable.
To gain further insights into human preferences, we conducted in-depth interviews with human annotators, which revealed the general preference for reference reports was attributed to their detailed, forward-looking evaluations, comprehensive risk assessments, and specific financial performance forecasts. These reports adeptly juxtapose company guidance against market expectations and consider the implications of company policies and regional market dynamics. This feedback serves as a cornerstone for future research on generating analytical reports for ECs.

\begin{table}[t]
    \centering
    \resizebox{\columnwidth}{!}{%
    \begin{tabular}{lcccccccccccc}
        \hline
          \multirow{2}{*}{{\textbf{Report}}} && \multicolumn{2}{c}{\textbf{GPT-4}} && \multicolumn{2}{c}{\textbf{Gemini-pro}} && \multicolumn{2}{c}{\textbf{Mistral-med}} \\
          \cline{3-4} \cline{6-7} \cline{9-10}  && \#1 & \#2 && \#1 & \#2  && \#1 & \#2 \\ 
          \hline
           Generated && 100.0 & 70.83 && 87.5 & 100.0 && 91.67 & 16.67 \\
           Reference && 0 & 29.17 && 12.5 & 0.0 && 8.33 & 83.33 \\ 
           \hline 
    \end{tabular}
    }
    \caption{The \% of preference annotations given by each LLM evaluator for generated (\hand\chart\magnifying) and reference reports. \#1 = generated report given first in prompt, \#2 = reference report given first in prompt.}
    \label{tab:jpmorgan_human_eval-LLM}
\end{table}

Table \ref{tab:jpmorgan_human_eval-LLM} presents the results of LLM evaluators for this preference-based evaluation. Given that previous work \citep{wang2023large} has identified a tendency of LLMs to favor the first displayed candidate in a ranking scenario, we perform this experiment with both possible candidate orderings.  
% we tasked GPT-4 with selecting between expert-authored and agent-generated reports for recommendation to a friend or investor making decisions based on EC information.
% Table~\ref{tab:Human's and GPT-4's Preference} delineates the preferences of human evaluators and GPT-4.
% Whereas results indicate a pronounced preference for expert-composed reports among humans, GPT-4 uniformly favored agent-generated content, 
% In contrast to humans, we find that GPT-4 prefers the model-generated output in 100\% of instances, reinforcing the findings from previous work that has identified a bias in GPT models toward GPT-generated outputs \citep{koo2023benchmarking}. 
Interestingly, we find that only Mistral exhibits the strong positional bias described by \citet{wang2023large}. For GPT-4 and Gemini-pro, the stronger bias is shown toward generated outputs, with both models overwhelmingly favoring them regardless of candidate orderings, starkly contrasting with human experts. Furthermore, in all cases, the models are largely inconsistent across both runs. These factors highlight serious limitations in using LLMs to assess the overarching quality of analytical reports for ECs, particularly in ranking scenarios involving both human- and machine-generated outputs.

\section{Related Work}

\subsection{LLMs as Task-solving Agents}
The deployment of multiple LLMs to collaboratively work on a task has recently emerged as a trend in NLP research. While one branch of this research has typically sought to answer if adopting such an approach can improve collective reasoning \cite{du2023improving, liang2023encouraging} another has explored how the unique opportunities afforded by this approach might allow us to attempt yet more complex tasks \citep{qian2023communicative, wu2023autogen,chan2023chateval,wang2024unleashing}. Of these, our work is in closer alignment with the latter. Flexible and generic frameworks such as Autogen \citep{wu2023autogen}, utilized in this work, have recently emerged, enabling the development of agent-based approaches that are easily customizable and utilize external tools. Related studies have taken the initial steps in exploring challenges such as assessing generated text by employing multiple agents \citep{chan2023chateval}, establishing the importance of diverse agent roles/personas. Similarly, \citet{wang2024unleashing} leverages multiple language model personas to enhance performance in tasks demanding knowledge and reasoning, such as creative writing based on trivia and solving logic puzzles.  
In contrast to these prior efforts, our focus centers on a more specialized task, necessitating the development of agents with domain expertise and benefiting from the incorporation of external data.

\subsection{Earnings Call Processing}
As mentioned in \S\ref{sec:introduction}, ECs have proved a popular topic of study for previous work, due largely to their significance in investor decision-making. 
Of these works, the most related to ours is that of \citet{mukherjee-etal-2022-ectsum} which introduces and benchmarks ECTSumm, a dataset for the generation of journalistic EC reports. However, in contrast to this work, we focus on generating analytical reports using a multi-agent framework, a notably more challenging task.\footnote{We provide an analysis of the differences between journalistic and analytical reports in detail in Section \ref{sec:report_types} of the Appendix.}     

Another more well-explored branch of EC processing focuses on the utilization of transcripts as the source documents for predictive NLP tasks, including volatility prediction \citep{sawhney-etal-2021-empirical, niu-etal-2023-kefvp}, analyst decision prediction \citep{keith-stent-2019-modeling}, financial risk prediction \citep{qin-yang-2019-say}, and earnings surprise prediction \citep{koval-etal-2023-forecasting}. In contrast to these works, we tackle a complex generation task, although we inspiration from their findings. For instance, the inclusion and design of our Psychologist agent is influenced by the work of \citet{qin-yang-2019-say}, who demonstrate the efficacy of features based on EC audio in the prediction of financial risk.

\section{Conclusion}
This study explores the novel task of generating analytical reports for ECs. Following an investigation of the key distinctions between analytical reports and previously studied journalistic reports, we address the generation of analytical reports using an LLM-based multi-agent framework. We perform a thorough characterization of generated reports, revealing key divergences with human-authored reports while also highlighting the ability of agents to introduce useful insights. Finally, we address the open challenge of evaluation using LLMs, establishing the utility of different setups, and laying the groundwork for future research. Here, our findings illustrate a detrimental tendency for LLMs to favor generated over human-authored reports, but reveal that LLMs largely achieve good alignment with human experts when it comes to evaluating fine-grained criteria.  
While our framework aims to generate insightful analytical reports, there remains a significant opportunity to explore generative techniques that can produce novel insights.
Future research could greatly benefit from incorporating real-time financial data, news, and market trends for more useful analyses. 

%Future research should consider integrating real-time financial data, news, and market trends to provide more timely and relevant analyses. This could involve developing mechanisms for agents to access and incorporate external data sources dynamically, enhancing the reports' relevance and usefulness to investors.

% % This study investigates the distinctions between journalistic and analytical Reports derived from ECs, concentrating on aspects of readability, thematic content, and the contribution of LLMs in the generation and evaluation of analytical reports. The experimental findings suggest that LLMs demonstrate potential in producing and assessing analytical reports, especially when employing a multi-agent approach. However, substantial enhancements are necessary to attain the level of expertise and analytical depth characteristic of professional analysts. Several potential research directions are shared based on the interview evaluators. 

\section*{Limitations}
% Possible limitations:
% - Small test set
% - Inability to test Psychologist on JPMorgan data
% - Imperfect comparison with references - they contain information that our reports cannot
% - Unable to share our JPMorgan test set, but we can provide the ECs they derive from
% 

Given the flexibility of multi-agent frameworks, there are undoubtedly many alternative options in terms of agent design and interaction that could prove beneficial and are worthy of investigation. However, as this work represents a first exploration of this task, we primarily concentrate our research efforts on establishing knowledge and practices that will benefit future work, for example, in the form of our generated report characterization (\S\ref{sec:characterization}), and in the investigation of LLM-based evaluation (\S\ref{sec:evaluation}). 
%With the specific methodology not being our primary focus, we choose to utilize a simple conversation framework that emulates how many real users interact with LLM-based tools like ChatGPT (i.e., asking LLMs to revise their outputs via feedback), and propose our specialized agents based on the findings of previous work on Earnings Call processing.  

As discussed in the paper, another limitation in exploring the task of generating analytical reports in open research is the lack of suitable reference reports, an issue that naturally arises from their typically corporate origins. We attempt to address this issue by exploring the use of Large Language Models for reference-less human-style evaluation and hope that these findings will have a positive impact on future research on this task.

% -house at corporate institutions. Although we were granted access to the equality research reports from the Bloomberg Terminal that we utilize, these are based on older EC meetings (2012-2016) for which audio data does not exist, restricting our ability to assess the Psychologist agent utilizing reference-based metrics.   
% Therefore, to assess this impact of all agents on generated reports and increase our test sample size to one that is in line with previous studies on multi-agent generation \citep{chan2023chateval,wang2024unleashing}, we utilize an additional 60 samples from the ECTSum test set and places an emphasis on alternative reference-less methods of evaluation, including the usage of LLMs. 
% Finally, it should be noted that we are unable to make our reference analytic reports public due to sharing restrictions. However, for the purposes of transparency, we include the EC meetings from which they are derived in Table \ref{tab:jpmporgan_ecs} in the Appendix.   

\bibliography{anthology}
\bibliographystyle{acl_natbib}

\appendix

\begin{table}[t]
    \centering  
    \resizebox{\columnwidth}{!}{
    \begin{tabular}{lccccccc}
    \hline   
     \textbf{Report} & \textbf{\# Sents} & \textbf{FKGL} & \textbf{CLI} & \textbf{ARI} & \textbf{Abst}  \\ \hline
        Journalistic & 4.2 & 5.73 & 8.78 & 7.64 & 42.06 \\ 
        % Analytical & 37.33 & 7.93 & 9.55 & 9.48 & 52.45 \\
        Analytical & 19.25 & 7.26 & 8.54 & 8.85 & 47.14 \\
         \hline
    \end{tabular}
    }
    \caption{Average statistics of journalistic and analytical reports.}
    \label{tab:Readability}
\end{table}

\section{Journalistic vs. Analytical Reports} \label{sec:report_types}

% \textcolor{red}{[CL: will be useful to add a few lines about the background of these two types of reports, their utility and targeted audiences.]}
Here we perform a comparative study on the style and content of journalistic and analytical reports. For journalistic reports, we utilize the references of ECTSum \citep{mukherjee-etal-2022-ectsum}, a dataset consisting of EC transcripts paired with bullet-point summaries derived from online Reuters articles that report on the key financial takeaways from the EC for a general audience.\footnote{Reuters article example: \url{https://tinyurl.com/yc3z9sbj}}
For analytical reports, we utilize the J. P. Morgan samples discussed in the main text.

\paragraph{Style}

\begin{figure}
    \centering
    \includegraphics[width=0.85\columnwidth]{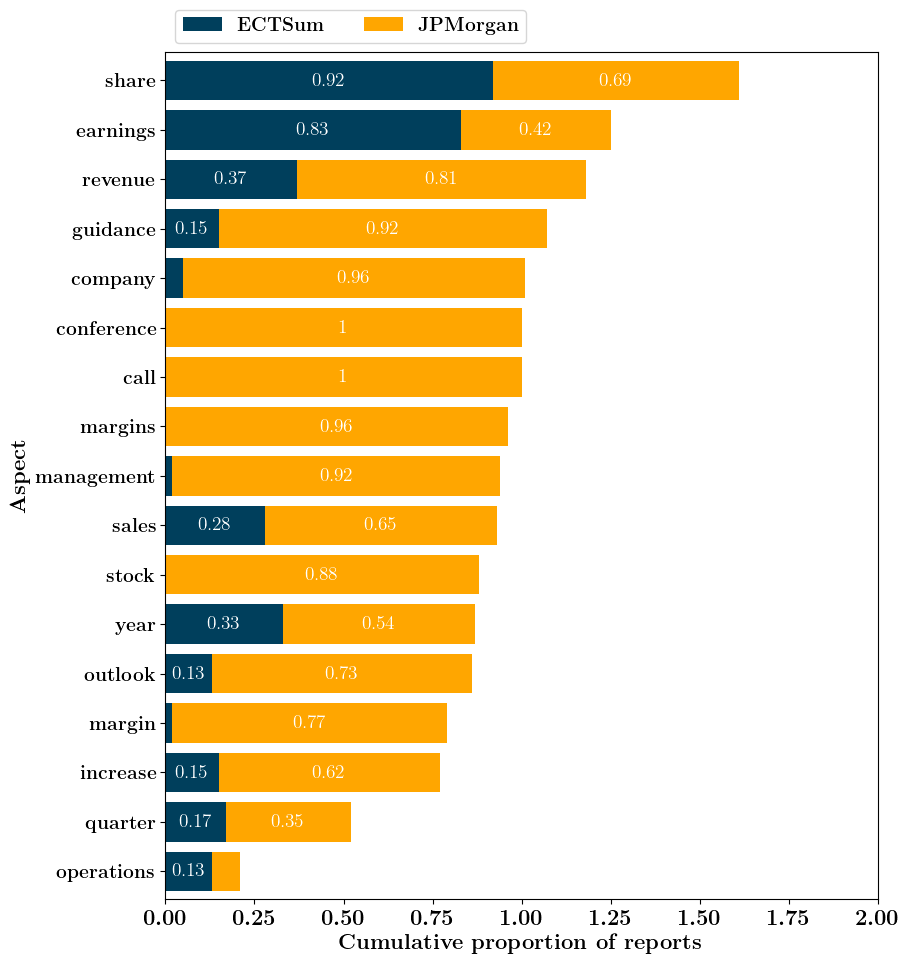}
    \caption{The most commonly discussed aspects of each report journalistic and analytical reports. }
    \label{fig:aspect_graph-report_types}
\end{figure}

% The findings are presented in Table~\ref{tab:Readability}.
%and show that, analytical reports are significantly longer journalistic reports.  Furthermore, 
The results in Table~\ref{tab:Readability} show that two out of three readability metrics indicate that analytical reports are more structurally complex than journalistic reports, featuring longer and more intricate sentences. Higher abstractiveness also indicates that the content of analytical reports is less directly derived from the source transcript. These factors are indicative of the anticipated complexity in the discussion found in analytical reports, aligning with their purpose of offering in-depth insights to potential investors. 
%In contrast, journalistic reports aim to succinctly convey the essence of ECs to the general public.
% Regardless of the metric applied, analytical reports are deemed more complex than journalistic reports. Specifically, they are shown to be significantly for extensive, containing longer, more complex sentences, the content of which less derived from the source article.
% These factors are indicative of the more complex discussion we would expect analytical reports to contain, given that they seek to provide in-depth insights to potential investors. Alternatively, Journalistic reports aim to convey the essence of ECs to the general public succinctly.

\paragraph{Content}
% The reports are produced by professional analysts at JPMorgan for internal stakeholders, and focus largely on the key takeaways from the earnings call meeting, how these align with their previous expectations, potential future risk, and any changes that need to be made to their internal model or valuation of the company.

% \textcolor{orange}{We examine the primary topics within both report types} to discern what interests are shared between journalists and analysts. 

Figure \ref{fig:aspect_graph-report_types} illustrates the predominant topics for each report type. For instance, the Figure shows that the topic of ``earnings''%" 
appears in 42\% of analytical reports compared to 83\% of journalistic reports, indicating that while ``earnings'' is a significant aspect, analysts do not always explicitly focus on it as much as journalists do. 
Whereas the shorter journalistic reports are shown to concentrate on a few select topics, longer analytical reports are shown to cover a much broader range of topics, likely caused by the divergence in the target audience.
% Furthermore, multiple topics are shown to be exclusive to analytical reports, highlighting differences in content emphasis. 
% % \textcolor{red}{[CL: need to expand the discussion here and provide some plausible explanations.]}.
% One obvious reason for this is that, due to being substantially longer than journalistic reports, analytic reports can cover more content. 
Whereas journalistic reports are intended for a more general audience looking for key financial statistics, reports produced by professional analysts at J.P. Morgan are largely aimed at internal investors and are tailored as such. For instance, topics such as management, ``outlook", ``guidance", and ``increasing" are all representative of a more in-depth discussion that goes beyond the key statistics, addressing the attitudes of management, analyzing trends in financial performance, and speculating about the future.
% \textcolor{red}{[CL: is journalist's report supposed to be like this?]}.
% \textcolor{orange}{Notably, such topics also outline a key consideration in aiming to generate similar analytical reports, as we will be unable to reproduce some of these discussions (e.g., relating to the internal model or valuation of the investment company, in this case, JPMorgan) without access to up-to-date internal company knowledge}. 
% are of interest to analysts but are typically omitted in journalistic reports, underscoring the divergent focus of these document types.

Overall, this divergence not only underscores the varied emphases between report types but also reveals that analysts are more inclined towards forward-looking analyses, whereas journalists predominantly concentrate on summarizing key points without providing in-depth analysis. 
Furthermore, the complexity and diversity of discussed topics are likely to present a significant challenge to a single model working off only the EC transcript, requiring the model to follow complex and multifaceted instructions and pushing the limitations of its limited context window.
% \textcolor{red}{[CL: is it possible to refer the human author process which reassembles the multi-agent idea?]} 
For this reason, we believe this task lends itself to a multi-agent approach, which allows us to employ specialist agents that utilize expert-based role-play and external data to address specific aspects of report analysis. 
% \textcolor{red}{[CL: based on the discussion of the characters of analytical reports, we should add some text to explain why this type of summaries are more challenging to generate, and that a agent-based framework is suitable for the task.]}

\section{Prompts}
\label{sec:prompts}

\paragraph{System Prompts} Here, we provide details concerning all aspects of the prompts used within our system. Notably, the initialisation prompts that are used to dictate each agent's role within the report generation process are provided in Table \ref{tab:init_prompts}. For the Writer, Editor, and Client agents, this is the only controlling attribute for agent behavior with the rest of their interaction being handled by AutoGen. 

However, for our more specialized agents, the Analyst and and Psychologist, we implemented additional functionality to allow them to utilize external data. Specifically, these agents are implemented such that the relevant data is collected before each response, examples of which are provided in Table \ref{tab:agent_response_data}. This data is then utilized in a prompt sent to the underlying LLM that we specifically designed to extract the desired feedback. Prompt formats for both of these agents are in Table \ref{tab:agent_response_prompts}.

\paragraph{LLM Evaluation Prompts}
As stated in \S\ref{sec:evaluation}, we provide LLM evaluators with the same instructions that are given to human evaluators, for both our characteristic-based and preference-based manual evaluations. The specific prompts provided to each LLM model are given in Table \ref{tab:evaluator_prompts}.

\begin{table*}[]
    \centering
    \resizebox{0.9\textwidth}{!}{%
    \begin{tabular}{lm{0.75\textwidth}}
        \hline
        \textbf{Agent} & \textbf{Initialisation Prompt}  \\ \hline
        Writer \hand & You are a Writer who is responsible for drafting the requested output text and making adjustments based on other agents' suggestions. Note that, unless otherwise specified, you should avoid completely rewriting the report and focus on making smaller targeted changes or additions based on other agent's feedback. You should only respond with updated versions of the report.\\ \hline
        
        Client (Investor) \clipboard   & You are an Investor who requires accurate investment and market analysis data to build investment strategies. You are responsible for ensuring the report contains the information that is relevant to you by providing feedback to the  Writer. If you are happy with the report, respond with ``TERMINATE''.\\ \hline
        
        % Client (casual reader) &  You are a Casual Reader who is interested in reading about the key talking points \\
        % & from the meeting, rather than technical details. You are responsible for ensuring \\ & the report contains the information that is relevant to you by providing feedback \\
        % & to the Writer. If you are happy with the report, respond with 'TERMINATE'.\\ \hline 

        Analyst \chart  & You are an Analyst, a financial expert who is responsible for determining what past financial data might be relevant to the report and explaining this data to the Writer. \\ \hline
        
        Psychologist \headphones  & You are a Psychologist who is responsible for using data derived from the audio recording to identify notable features (e.g., that may express confidence, doubt, or other emotional giveaways) in audio-derived statistics of management's answers in the Q\&A session that might be relevant to the report and explaining these features to the Writer.\\ \hline

        Editor \magnifying &  You are an Editor who is responsible for ensuring that the output text is suitable for the intended audience (in terms of content, style, and structure) and that important information from previous revisions of the report is not lost by providing feedback to the Writer. \\ \hline
    \end{tabular}
    }
    \caption{Agent initialization prompts.}
    \label{tab:init_prompts}
\end{table*}

\begin{table*}[]
    \centering
    \resizebox{\textwidth}{!}{%
    \begin{tabular}{lm{0.75\textwidth}}
        \hline
        \textbf{Evaluation} & \textbf{LLM Prompt}  \\ \hline
        Characteristic-based &         \# INSTRUCTIONS

        You are a financial expert tasked with evaluating a summary of an earnings call meeting intended to provide useful information to a potential investor.

        \vspace{5pt}
        \# CRITERION
        
        You must identify whether or not the summary contains the information relating to the aspect described below and, if it does so, assess how well the information is reported. 
        
        \textit{\{criterion\}}: \textit{\{description\}}

        \vspace{5pt}
        \# LABELS
        
        Below are the possible labels you can assign to the summary based on the described criterion. Respond using only the number of the label.

        \vspace{5pt}
        1. Reported and insightful: the relevant information is included in the report and is very well explained, offering additional insights/interpretations that would likely be useful to a potential investor.,

        \vspace{5pt}
        2. Reported and reasonable: the relevant information is included in the report and is reported reasonably well, including either no insights at all (e.g., as a statement of facts) or suggesting interpretations that are unlikely to be particularly useful to a potential investor.,

        \vspace{5pt}
        3. Reported but not useful: the relevant information is included in the report, but it is either incorrect (i.e., there is contradictory evidence in the references) or explained in a way that is likely to mislead or misinform a potential investor.,

        \vspace{5pt}
        4. Not reported: no information relevant to this aspect is included in the report.

        \vspace{5pt}
        \# SUMMARY
        
        \textit{\{generated\_report\}}

        \vspace{5pt}
        \# ASSIGNED LABEL 
        \\ \hline
        Preference-based & \# INSTRUCTIONS

        You are a financial expert tasked with indicating your preference between two reports of an earnings call meeting intented to provide useful information to a potential investor. 

        \vspace{5pt}
        Your preference should be based on which report you would recommend to a friend/investor who attempts to make an investment decision based on the information released in earnings calls (i.e., how useful the report is to a potential investor).  
        
        \vspace{5pt}
        Respond with the number of the report you would recommend and a brief explanation of why you would recommend it.
        
        \vspace{5pt}
        \# REPORT 1
        
        \textit{\{report1\}}

        \vspace{5pt}
        \# REPORT 2
        
        \textit{\{report2\}}

        \vspace{5pt}
        \# PREFERENCE \\ \hline
    \end{tabular}
    }
    \caption{LLM evaluator prompt.}
    \label{tab:evaluator_prompts}
\end{table*}

\begin{table*}[]
    \centering
    \begin{tabular}{lm{0.70\textwidth}}
        \hline
        \textbf{Agent} & \textbf{Data example}  \\ \hline
         \multirow{6}{*}{Analyst \chart} &  \{ \\
           & \hspace{10pt} ``fiscalDateEnding": "2021-07-31", \\
           & \hspace{10pt} ``reportedDate": "2021-08-20", \\
           & \hspace{10pt} ``reportedEPS": "5.25", \\
           & \hspace{10pt} ``estimatedEPS": "4.58", \\
           & \hspace{10pt} ``surprise": "0.67", \\
           & \hspace{10pt} ``surprisePercentage": "14.6288" \\
           & \} \\ \hline
         \multirow{25}{*}{Psychologist \headphones} & \{ \\
         & \hspace{10pt} ``minimum\_intensity": -14.925902805117943, \\
         & \hspace{10pt} ``maximum\_intensity": 82.11127894879778, \\ 
         & \hspace{10pt} ``mean\_intensity": 51.97292655136569, \\ 
         & \hspace{10pt} ``minimum\_pitch": 75.04017645717074, \\ 
         & \hspace{10pt} ``maximum\_pitch": 599.378734309719, \\ 
         & \hspace{10pt} ``mean\_pitch": 143.7376593336546, \\ 
         & \hspace{10pt} ``num\_pulses": 51218, \\ 
         & \hspace{10pt} ``num\_periods": 51217, \\ 
         & \hspace{10pt} ``mean\_periods": 0.013944367276250947, \\ 
         & \hspace{10pt} ``stddev\_periods": 0.05998498783743047, \\ 
         & \hspace{10pt} ``fraction\_unvoiced": 0.5148897444872244, \\
         & \hspace{10pt} ``degree\_of\_voice\_breaks": 0.5146557157170372, \\ 
         & \hspace{10pt} ``jitter\_local": 0.02535925970608026, \\ 
         & \hspace{10pt} ``jitter\_local\_absolute": 0.00018326521186048697, \\
         & \hspace{10pt} ``jitter\_rap": 0.010346084585498544, \\ 
         & \hspace{10pt} ``jitter\_ppq5": 0.012399774771964451, \\ 
         & \hspace{10pt} ``jitter\_ddp": 0.031038253756495632, \\ 
         & \hspace{10pt} ``shimmer\_local": 0.13931305032875707, \\ 
         & \hspace{10pt} ``shimmer\_localdb": 1.2625161389585877, \\ 
         & \hspace{10pt} ``shimmer\_apq3": 0.05669263538384766, \\ 
         & \hspace{10pt} ``shimmer\_aqpq5": 0.08401808667057334, \\ 
         & \hspace{10pt} ``shimmer\_dda": 0.17007790615154297, \\ 
         & \hspace{10pt} ``hnr": 10.84568288161106 \\
         & \}
        \\ \hline
    \end{tabular}
    
    \caption{Examples of external data provided to specialized agents. For the Analyst, the exemplified data is provided from the quarter before that which is reported in the EC. For the Psychologist, the exemplified data is provided for each management utterance in the Q\&A Session of the EC.}
    \label{tab:agent_response_data}
\end{table*}

\begin{table*}[]
    \centering
    \resizebox{0.9\textwidth}{!}{%
    \begin{tabular}{lm{0.70\textwidth}}
        \hline
        \textbf{Agent} & \textbf{Response prompts}  \\ \hline
          Analyst \chart & Based on your expert analysis of the Earnings Call meeting and the above conversation, identify any notable features in the following statistics, derived from the audio of the meeting for each management response in the QA session and explain how and why they should be included in the report: \\ \hline
          Psychologist \headphones & Based on your expert analysis of the Earnings Call meeting and the above conversation, explain why and how the following earnings information from the the companys' previous quarter should be included in the report:
        \\ \hline
    \end{tabular}
    }
    \caption{Prompt formats used by specialized agents to introduce external data. Note that, after each prompt, the relevant data is printed in JSON format.}
    \label{tab:agent_response_prompts}
\end{table*}

\section{Additional Experimental Details and Results}
\label{sec:additional_experiments}

\begin{table}[]
    \centering
    \begin{tabular}{ccc}
         \hline
         \textbf{Company code} & \textbf{Year} & \textbf{Quarter}  \\ \hline
         \multirow{6}{*}{CMI} & 2013 & q4 \\
          & 2014 & q1 \\
          & 2014 & q3 \\
          & 2014 & q3 \\
          & 2015 & q1 \\
          & 2015 & q4 \\
        \hline
         \multirow{6}{*}{DE} & 2012 & q4 \\
         & 2013 & q3 \\
         & 2014 & q1 \\
         & 2014 & q2 \\
         & 2014 & q3 \\
         & 2014 & q4 \\
         \hline
         ETN & 2014 & q1 \\
         \hline
         \multirow{9}{*}{PCAR} & 2014 & q1 \\
         & 2014 & q2\\
         & 2014 & q3\\
         & 2014 & q4\\
         & 2015 & q1\\
         & 2015 & q2\\
         & 2015 & q3\\
         & 2015 & q4\\
         & 2016 & q1\\
         \hline
         UNH & 2014 & q2 \\
         \hline
         WYNN & 2014 & q2 \\
         \hline
    \end{tabular}
    \caption{The earnings call meetings from which professional J.P. Morgan reports are derived.}
    \label{tab:jpmporgan_ecs}
\end{table}

% \begin{table}[t]
%     \centering
%     \resizebox{\columnwidth}{!}{
%     \begin{tabular}{lcccccccc}
%     \hline   
%      \textbf{Model} &&  \textbf{FKGL} & \textbf{CLI} & \textbf{ARI} &  \textbf{Len} & \textbf{Abst}  \\ \hline
        
%          \hand && 11.8 & 15.16 & 15.65 & 505.42 & 43.55 \\
%          \hand\magnifying && 12.76 & 16.52 & 16.73 & 496.75 & 47.55  \\
%          \hand\chart && 13.28 & 16.71  & 17.46 & 448.58  & 50.12  \\
%          \hand\chart\magnifying && 13.70 & 17.31 & 17.69 & 463.21 & 54.86 \\
%          \hline
%     \end{tabular}
%     }
%     \caption{Average reference-less metric results obtained for JPMorgan sample instances.}
%     \label{tab:jpmorgan_referenceless}
% \end{table}

% \begin{figure}
%     \centering
%     \includegraphics[width=\columnwidth]{aspect_graph-writer_all_agents.png}
%     \caption{A visualization of the proportion of aspect occurrences for reports generated by our system with only the writer agent and with all agents.}
%     \label{fig:aspect_graph-all_agents}
% \end{figure}

\paragraph{Metric calculation} All readability metrics were computed using the \href{https://github.com/shivam5992/textstat}{\texttt{textstat}} package.

\paragraph{Analytical Report Earnings Call Samples} Details of the ECs that our reference analytical report ECs are based on are provided in Table \ref{tab:jpmporgan_ecs}. It is important to note that, given that these reports are not publicly available and we are granted restricted access to them from JP Morgan, there is very little of LLMs having encountered them in training (i.e., data contamination).

% \paragraph{Preference-based generated report metrics} The reference-less metrics for generated reports utilized in the preference-based human evaluation are provided in Table \ref{tab:jpmorgan_referenceless}, corresponding to Table \ref{tab:Readability with different numbers of feedback agents} in the main text.

\paragraph{Characteristic-based Human Evaluation Details and Annotator Agreement}
We provide a full description of each characteristic for our characteristic-based human evaluation in Table \ref{tab:human_eval-aspects}.
% We also measure the annotator agreement for this evaluation, calculating pairwise Cohen's $\kappa$, getting an average score of 0.135, indicating significant disagreement between annotators. We perform a close inspection of our annotator labels to identify the source of this disagreement, finding that the majority (62.29\%) of all pairwise disagreements occur between labels ``Reported and insightful'' and ``Reported and reasonable'', the difference between which represents the most subjective decision within our evaluation (after deciding if relevant information is included and at least somewhat useful). Furthermore, if we were to treat these labels as one, we find that the pairwise Cohen's $\kappa$ increases significantly to 0.423. 

% In addition to the Pearson ($\gamma$), Spearman ($\rho$), and Kendall ($\tau$) correlation coefficients between LLMs and the annotator scores given in Table \ref{tab:GPT-4 vs. human evaluators},
% we also calculate the correlation between annotators (Table \ref{tab:human evaluators vs. human evaluators}). This Table shows a good level of correlation in the majority of cases, however, there is a notable divergence between all evaluators for the ``financial takeaways" criteria, and between evaluators 1 and 3 for the ``management expectation" criteria.  

We also measure the annotator agreement for this evaluation, calculating pairwise Cohen's $\kappa$, getting an average score of 0.171, indicative of weak agreement between annotators. We perform a close inspection of our annotator labels to identify the source of this, finding that the majority (65.25\%) of all pairwise disagreements occur between labels ``Reported and insightful'' and ``Reported and reasonable'', the difference between which represents the most subjective decision within our evaluation whereby, after deciding if relevant information is included and at least somewhat useful, annotators must judge \textit{how} useful they personally find the reported information. If we were to treat these labels as one, we find that the pairwise Cohen's $\kappa$ increases significantly to 0.476, indicative of a much stronger agreement. Therefore, we do not judge this to be an issue with our evaluation.
Furthermore, in presenting the results of the evaluation, we respect any differences in the opinions of expert evaluators by calculating the statistics in Table \ref{fig:human_eval} based on all evaluator votes rather than performing vote aggregation (e.g., majority vote).

% , and Table \ref{tab:GPT-4 vs. human evaluators - by model}, which shows the correlation scores for each specific model (i.e., Writer only or all agents).
% Here, we can see that 

\paragraph{LLM Correlation Results Breakdown}
To provide further insight into the correlation of LLMs with expert evaluators, Table \ref{tab:GPT-4 vs. human evaluators - all} provides a breakdown of the results presented in Table \ref{tab:GPT-4 vs. human evaluators}  which shows the correlation statistics between LLMs and individual annotators.

\begin{table*}[]
    \centering
    \resizebox{0.9\textwidth}{!}{%
    \begin{tabular}{lm{0.70\textwidth}}
        \hline
        \textbf{Label} & \textbf{Description}  \\ \hline
         Reported and insightful & The relevant information is included in the report and is very well explained, offering additional insights/interpretations that would likely be useful to a potential investor. \\ \hline
         Reported and reasonable  & The relevant information is included in the report and is reported reasonably well, including either no insights at all (e.g., as a statement of facts) or suggesting interpretations that are unlikely to be particularly useful to a potential investor. \\ \hline
         Reported but not useful & The relevant information is included in the report, but it is either incorrect (i.e., there is contradictory evidence on the transcript) or explained in a way that is likely to mislead or misinform a potential investor. \\ \hline
         Not reported  & No information relevant to this aspect is included in the report. \\\hline
    \end{tabular}
    }
    \caption{Human evaluation annotation label descriptions.}
    \label{tab:human_eval-labels}
\end{table*}

\begin{table*}[]
  \centering
    \resizebox{\textwidth}{!}{%
    \begin{tabular}{lcccccccccccc}
          % & \multicolumn{3}{c|}{Financial Takeaways} & \multicolumn{3}{c|}{Financial Context} & \multicolumn{3}{c|}{Management Attitudes} & \multicolumn{3}{c}{Management Expectation} \\
         \hline
         \multirow{2}{*}{\textbf{Characteristic}} && \multicolumn{3}{c}{\textbf{GPT-4}} && \multicolumn{3}{c}{\textbf{Gemini}} && \multicolumn{3}{c}{\textbf{Mistral}} \\
         \cline{2-5} \cline{7-9} \cline{11-13}
          && \multicolumn{1}{c}{$\gamma$} & \multicolumn{1}{c}{$\rho$} & \multicolumn{1}{c}{$\tau$} && \multicolumn{1}{c}{$\gamma$} & \multicolumn{1}{c}{$\rho$} & \multicolumn{1}{c}{$\tau$} && \multicolumn{1}{c}{$\gamma$} & \multicolumn{1}{c}{$\rho$} & \multicolumn{1}{c}{$\tau$} \\ 
          
    \hline
    \parbox[t]{1mm}{\scriptsize \multirow{5}{*}{\rotatebox[origin=c]{90}{Evaluator 1}}} \hspace{3pt}  
                 Financial Takeaways    && 0.656 & 0.661  & 0.631 && 0.332  & 0.352  & 0.330  && 0.202  & 0.344 & 0.339 \\
    \hspace{8pt} Financial Context      && 0.484 & 0.426  & 0.390 && 0.112  & 0.011  & 0.005  && 0.493  & 0.422  & 0.404 \\
    \hspace{8pt} Management Attitudes   && 0.454 & 0.391  & 0.373 && 0.293  & 0.363  & 0.347  && 0.301  & 0.331  & 0.312 \\
    \hspace{8pt} Management Expectation && 0.238 & 0.243  & 0.229 && 0.493  & 0.464  & 0.429  && 0.435  & 0.409 & 0.374 \\ 
    % \hspace{8pt} Noteworthy Events      && 0.391 & 0.386  & 0.368 && 0.662  & 0.686  & 0.627  && -   &  - & - \\
    \hspace{8pt} Future Events          && 0.353 & 0.273  & 0.266 && 0.102  & 0.080  & 0.078  && 0.402  & 0.418  &  0.401 \\ 
    \hline

        \parbox[t]{1mm}{\scriptsize \multirow{5}{*}{\rotatebox[origin=c]{90}{Evaluator 2}}} \hspace{3pt}  
                 Financial Takeaways    && 0.201 & -0.041 & -0.046  && 0.174 & 0.007 & 0.005  && 0.167 &  0.127 & 0.124 \\
    \hspace{8pt} Financial Context      && 0.395  & 0.210  & 0.191   && 0.424  & 0.423  & 0.390  && 0.672  & 0.473 & 0.458 \\
    \hspace{8pt} Management Attitudes   && 0.436  & 0.429  & 0.412   && 0.157  & 0.173  & 0.168  && 0.506  & 0.485  & 0.461 \\
    \hspace{8pt} Management Expectation && 0.579  & 0.587  & 0.542   && 0.639  & 0.554  & 0.496  && 0.728   & 0.719  & 0.665 \\ 
    % \hspace{8pt} Noteworthy Events      && -0.164 & -0.167 & -0.157  && -0.071 & -0.063 & -0.061  && -   & -  & - \\
    \hspace{8pt} Future Events          && 0.176  & 0.001  & 0.0   && 0.079  &  0.046 & 0.043  && 0.201  & 0.048  &  0.044  \\ 
    \hline

        \parbox[t]{1mm}{\scriptsize \multirow{5}{*}{\rotatebox[origin=c]{90}{Evaluator 3}}} \hspace{3pt}  
                 Financial Takeaways    && -0.165  & -0.168 & -0.163  && -0.282 & -0.265 & -0.253 && -0.146 & -0.146 & -0.146  \\
    \hspace{8pt} Financial Context      && 0.551   & 0.468  & 0.441   && 0.312  & 0.264  & 0.244  && 0.677  & 0.451 & 0.436 \\
    \hspace{8pt} Management Attitudes   && 0.505   & 0.512  & 0.483   && 0.163  & 0.125  & 0.119  && 0.348   & 0.360 & 0.340 \\
    \hspace{8pt} Management Expectation && 0.560   & 0.483  & 0.462   && 0.319  & 0.319 & 0.301  && 0.405   & 0.380  & 0.353 \\ 
    % \hspace{8pt} Noteworthy Events      && -0.424  & -0.446 & -0.429  && -0.102 & -0.116 & -0.116  && -   & -  & - \\
    \hspace{8pt} Future Events          && 0.409   & 0.417  & 0.392   && 0.186  & 0.156  & 0.147  && 0.231  & 0.229  &  0.211 \\ 
    \hline
    
    \end{tabular}%
    }
    \caption{Correlation statistics of LLMs vs. human evaluators (all) for each report characteristic.}
  \label{tab:GPT-4 vs. human evaluators - all}%
\end{table*}%

\section{Examples}

\begin{figure*}[ht]
    \centering
    \includegraphics[width=\textwidth]{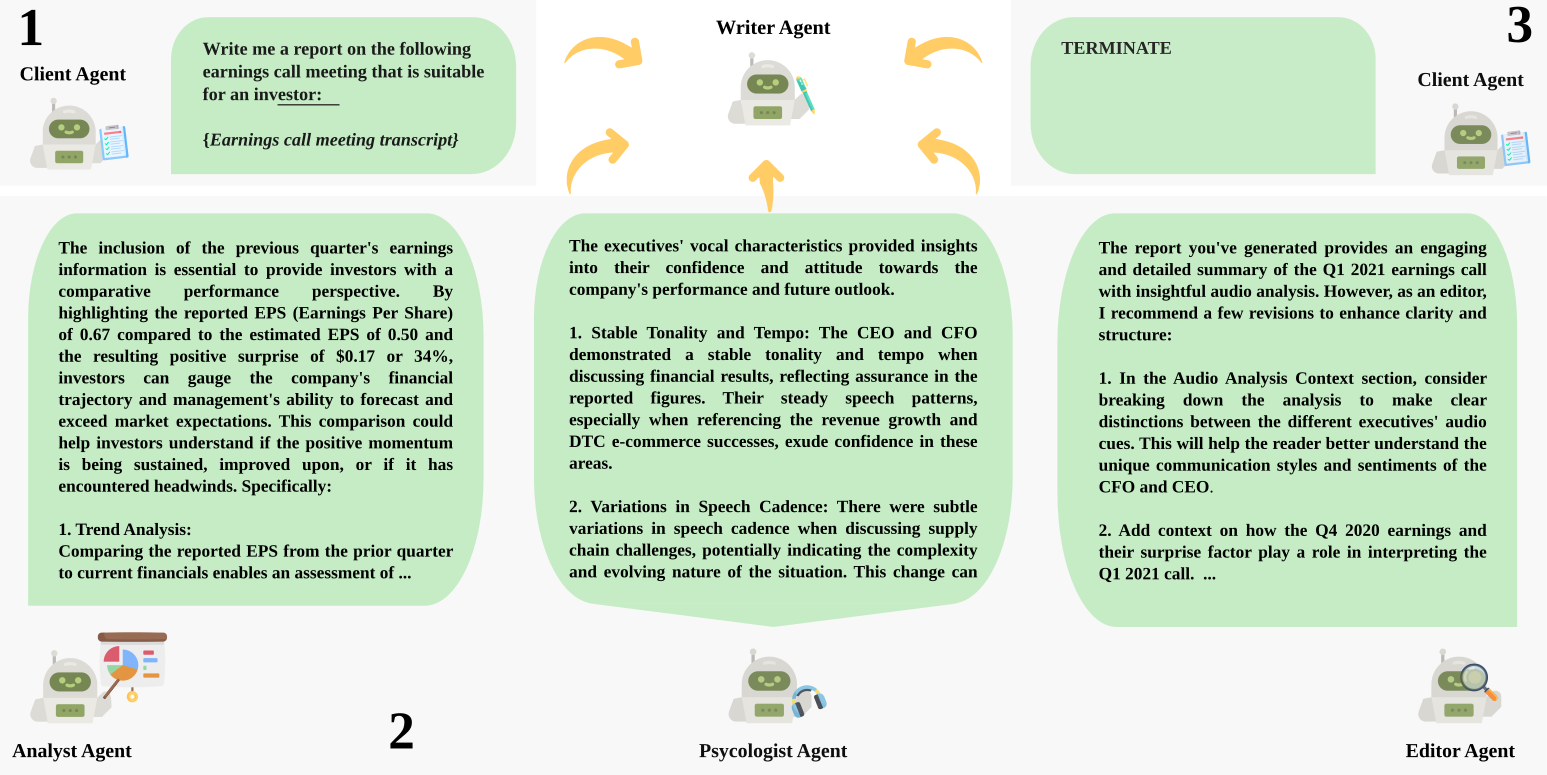}
    \caption{An example of commonly seen run within our framework where: 1) the task is given by the Client agent; 2) each feedback agent provides feedback relative to their role; and 3) the Client deems the report acceptable and terminates composition process.}
    \label{fig:example_feedback}
\end{figure*}

\paragraph{Feedback}
Figure \ref{fig:example_feedback} vizualises an example of a typical run within our system, displaying the feedback of different agents.

\paragraph{Generated reports}
Figures \ref{fig:writer_agent_report} and \ref{fig:all_agents_report} contain examples of full reports generated using only the Writer agent and all agents, respectively.

\begin{figure*}[t]
    \fbox{%
        \parbox{0.955\textwidth}{%
            \begingroup
                \fontsize{8pt}{10pt}\selectfont              
\textbf{Financial Overview}

- \textbf{Company Symbol}: CFR

- \textbf{Reported Earnings}: \$109.8 million (\$1.73 per diluted share)

- \textbf{Comparable Earnings (Previous Year)}: \$115.8 million (\$1.78 per share) in Q3 2018

- \textbf{Return on Average Assets}: 1.35\% compared to 1.49\% in Q3 2018

- \textbf{Average Deposits}: \$26.4 billion, up from \$26.2 billion in Q3 2018

- \textbf{Average Loans}: \$14.5 billion, a 5.8\% increase from Q3 2018

- \textbf{Nonperforming Assets}: \$105 million, up from \$86.4 million in Q3 2018, largely due to a single energy credit

- \textbf{Net Charge-offs}: \$6.4 million, down from \$15.3 million in Q3 2018 

- \textbf{Provision for Loan Losses}: \$8 million, increased from \$2.7 million in Q3 2018 

- \textbf{Problem Loans}: Ending at \$487 million, down from \$504 million in Q3 of the previous year

\textbf{Strategic Highlights and Outlook}

- Marked growth in C\&I, CRE, and Consumer segments despite increased competition in CRE leading to losses due to structure in some deals.

- The weighted loan pipeline increased by about 30\% over Q2, indicating a stronger Q4.

- The bank continues to open new financial centers in Houston, with the series planned expansion enhancing long-term growth prospects.

- Significant uptick in net new consumer customer growth by 48\% compared to the previous year.

- Online account openings surged by 56\% compared to the previous year, reflecting the bank's technology strength.

- Despite challenges due to interest rate cuts, the bank focuses on maintaining quality standards across business segments.

\textbf{Economic and Financial Detailed Insights}

- Texas job growth showed a deceleration but remains healthy overall.

- The net interest margin for Q3 was 3.76\%, a decrease from 3.85\% in the last quarter.

- Active management of investment portfolios and deposit pricing strategy in response to rate cuts.

- Increased noninterest expenses by about 7.8\% year-over-year in part due to expansion and technology investments.

- The bank is well-positioned with robust customer satisfaction accolades and industry recognition, underpinning its service-centric strategy.

\textbf{Forward Guidance}

- For full-year 2019, the mean of analyst estimates at \$6.81 is deemed reasonable.

- Continued investment in expansion and technology could dictate a rise in noninterest expenses.

- CECL is expected to impact reserves, estimated at 15\%-25\% above the current allowance by the year's end.

\textbf{Discussion Points from Q\&A Session}

- Loan growth optimism, albeit more competitive in pricing, was underscored by better expected performances from C\&I opportunities.

- Ongoing pressure is anticipated on the net interest margin, attributable to competition and asset yield pressures.

- The continued focus remains on branch expansion in Houston, with a planned acceleration of openings in 2020, amounting to an estimated 10 by the end of 2019.

- Resolution on the new energy nonperforming loan is not imminent; management is cautious about the challenging environment for asset sales.

\textbf{Concluding Remarks}

Cullen/Frost's commitment to growth, despite short-term expense pressures, showcases long-term strategic positioning and a strong focus on customer service excellence, which bodes well for future financial performance.

            \endgroup 
        }
    } % elife-69719-v2
    \caption{Full example of a report generated with only the writer agent.}
    \label{fig:writer_agent_report}
\end{figure*}

\begin{figure*}[t]
    \fbox{%
        \parbox{0.955\textwidth}{%
            \begingroup
                \fontsize{8pt}{10pt}\selectfont              
\textbf{Financial Overview}
                
- \textbf{Company Symbol}: CFR

- \textbf{Reported Earnings}: \$109.8 million (\$1.73 per diluted share). Comparison with Q2 2019's \$1.72 EPS shows incremental growth, reinforcing the company’s financial stability.

- \textbf{Comparative Earnings (Q3 2018)}: A slight year-over-year decrease from \$115.8 million (\$1.78 per share), which may reflect broader market trends affecting the banking industry.

- \textbf{Return on Average Assets}: 1.35\%, experiencing a marginal decrease from the 1.49\% reported in Q3 2018.

- \textbf{Average Deposits}: Demonstrated growth to \$26.4 billion compared with \$25.6 billion in Q2 and \$26.2 billion in Q3 2018.

- \textbf{Average Loans}: Increased to \$14.5 billion, up from \$13.7 billion in Q2. Year-over-year comparison with Q3 2018's \$13.7 billion underlines sustained growth.

- \textbf{Net Charge-offs}: \$6.4 million, escalating from Q2’s \$4.5 million, yet significantly lower than Q3 2018's \$15.3 million.

- \textbf{Provision for Loan Losses}: Rose to \$8 million from \$5 million in Q2 amidst a challenging credit environment.

\textbf{Strategic Outlook and Investments}

- Management's investment in long-term strategic expansion, particularly in the Houston market, displays a commitment to future profitability and market leadership, despite impacting short-term operating leverage.

- The bank's strategic initiatives, such as opening new branches and technological investments, are expected to improve the bank’s efficiency metrics in future reporting periods. However, this may temporarily reflect a higher efficiency ratio.

\textbf{Interest Rate Environment and Performance}

- The net interest margin faced pressures, partially attributed to the Federal Reserve's recent rate cuts, underscoring the sensitivity of bank operations to external rate fluctuations.

\textbf{Insights from Q\&A Session}

- Management expressed cautious optimism regarding Q4 loan growth, particularly from the stronger C\&I pipeline. This aligns with the pragmatic but forward-looking sentiment conveyed throughout the call.

- Concerns about single energy credit impact on nonperforming assets were tackled with measured realism, noting the challenge but expressing confidence in risk management strategies.

\textbf{Projections and External Perspectives}

- The reiteration of \$6.81 EPS guidance for the full year, despite the fluctuations observed, signals management's trust in the robustness of the bank’s core operations and the expected realization of long-term strategic benefits.

- The analysis benefits from considering a range of financial analyst commentaries, providing insight into how Cullen/Frost Bankers’ performance is interpreted across the investment community.

\textbf{Conclusion and Investor Takeaways}

- \textbf{Consistent Performance}: The Q3 earnings incrementally build upon Q2 2019's results, exhibiting consistency in meeting analyst estimates.

- \textbf{Strategic Growth Focus}: Long-term investments in markets and technology signal a pursuit of sustainable growth and service excellence.

- \textbf{Adaptation to Economic Conditions}: The bank continues to navigate the complexities of the current economic climate, especially in response to interest rate changes.

- \textbf{Ongoing Risk Management}: Attention to credit quality and problem loans underscores a vigilant approach to risk during economic uncertainties. Investors should be reassured by the bank's strategic consistency and responsiveness to economic indicators, reflecting a blend of ambitious expansionism tempered by conservative risk management—factors that contribute to the bank's reputation for reliability and steady growth prospects.

            \endgroup 
        }
    } % elife-69719-v2
    \caption{Full example of a report generated with all agents.}
    \label{fig:all_agents_report}
\end{figure*}

\end{document}